\documentclass[11pt]{article}
\usepackage{amssymb,amsmath,cite,times}
\usepackage{graphicx}

\newcommand{\bg}[1]{\boldsymbol{#1}}
\newcommand{\bm}[1]{\mathbf{#1}} 
\begin{document}

\title{\textbf{A Multicomponent Approach to Nonrigid Registration of
Diffusion Tensor Images}}
\author{Mohammed Khader and A. Ben Hamza\\
Concordia Institute for Information Systems Engineering\\
Concordia University, Montreal, QC, Canada}
\date{}
\maketitle
\begin{abstract}
We propose a nonrigid registration approach for diffusion tensor images using a multicomponent information-theoretic measure. Explicit orientation optimization is enabled by incorporating tensor reorientation, which is necessary for wrapping diffusion tensor images. Experimental results on diffusion tensor images indicate the feasibility of the proposed approach and a much better performance compared to the affine registration method based on mutual information in terms of registration accuracy in the presence of geometric distortion.
\end{abstract}
\medskip
\noindent{{\bf Keywords}:\, Diffusion tensor imaging; image registration; nonrigid; Tsallis entropy.

\newpage
\section{Introduction}
Diffusion tensor imaging (DTI) is a state-of-the-art magnetic resonance imaging (MRI) technique for analyzing the underlying white matter (WM) structure of the brain and investigating the microstructure of biological tissue, especially in the presence of fibrous structures~\cite{Basser:94}. At each voxel of a diffusion tensor (DT) image, the water diffusion anisotropy and preferred orientation can be measured and represented by a symmetric second-order tensor. The orientation of the resulting DT field represents the orientation of fiber bundles, and hence DTI is considered an ideal choice for studying and inspecting white matter metabolism in the brain. By detecting the orientation of water molecules in WM, DTI enables studying WM alteration across populations and provides a helpful tool for brain growth research~\cite{Filippi:05}. An important prerequisite for these studies is nonrigid image registration, which refers to the process of aligning two or more images of the same scene that were subject to elastic or nonrigid transformations so that their details overlap accurately. Extending nonrigid image registration from scalar images to DT images is, however, a challenging task, not only because of the multi-dimensionality of DT images, but also due in large part to the requirement of keeping DT orientation consistent with the anatomy after image transformation~\cite{Alexander:01}.

In recent years, a wide range of techniques have been proposed in the literature to tackle the nonrigid registration problem of DT images. The vast majority of these methods can be broadly classified into three main categories. The techniques in the first category ignore the orientation components of images and register scalar images associated with DTI data sets, such as the non-diffusion weighted images, MR-T2-weighted images, and fractional anisotropy (FA) maps~\cite{Jones:02,Guimond:02,Xu:03}. In the second category, the methods register actual tensor images without reorienting the tensors during registration~\cite{Kikinis:02,Alexander:99}. Ruiz-Alzola {\em et al.}~\cite{Kikinis:02} proposed a unified framework for nonrigid registration of scalar, vectorial and tensorial medical data. The framework measures image correspondence based on DT data by optimizing affine transformations in a certain restricted window of the image domain. Alexander {\em et al.}~\cite{Alexander:99} presented a multiresolution elastic matching method and proposed a similarity measure that combines DT and T1-weighted structural information by averaging their individual similarities. In all the aforementioned techniques, no tensor reorientation was applied during the registration and hence producing inaccurate image matching results. Techniques in the last category either explicitly optimize tensor reorientation~\cite{Zhang:06,Younes:06} or perform tensor reorientation after application of the final transformations; and hence no tensor reorientation is applied during the optimization step~\cite{Hecke:07}. Zhang {\em et al.}~\cite{Zhang:06} proposed a piecewise affine registration algorithm that incorporates DT data in the similarity measure in an effort to explicitly optimize tensor reorientation. In~\cite{Hecke:07}, Hecke {\em et al.} proposed a nonrigid coregistration algorithm based on a viscous fluid model, in which the quality of image matching is measured by the mutual information similarity measure. The tensor reorientation in this method is only carried out after the application of the final deformation field.

A number of DTI registration methods align T1- or T2- weighted images that are taken at the same time as DTI data sets, followed by applying the resulting deformation to DT images. T1- and T2-weighted images represent the WM structure as low-contrast regions, and hence registration based on these images poorly align the structure and orientation of the WM regions~\cite{Westin:03}. To circumvent this limitation and provide a more structural information, DT features are usually used. One possible feature that contains a high WM contrast is the scalar FA map, which has proven to be a suitable feature~\cite{Jones:02}. Guimond {\em et al.}~\cite{Guimond:02} proposed a multicomponent registration method based on eigenvalue images. Another feature that enhanced the quality of DTI registration is the DT components as reported in~\cite{Westin:03,Thirion:98}. Thirion {\em et al.}~\cite{Thirion:98} proposed a demons-based registration algorithm and used the sum of square differences as a similarity criterion based on DT elements. Alexander {\em et al.}~\cite{Alexander:01} reported that only rigid transformation should reorient the tensors to keep them consistent with anatomical structure of the image. For scalar measures such as the eigenvalues and the FA map, tensor reorientation is not required during registration due to the invariance to rigid transformations of their corresponding tensors. In contrast to FA, the DT elements contain orientation information, and hence the voxel intensities of the DT elements may have different values for a particular WM tract follows a different path in two subjects, where the FA can be similar. Because the intensity variation in the corresponding voxels has a local, spatial dependent nature, a DTI registration algorithm needs to accommodate both the alignment of intersubject images and the presence of nonlinear intervoxel intensity differences~\cite{Hecke:07}. Moreover, the widely used sum of square differences similarity measure assumes similar voxel intensity values in different images that only differ from each other by a Gaussian noise term. But the FA or eigenvalue image data are known to be non-Gaussian distributed due to nonlinearity in the calculation of the eigenvalue system~\cite{Anderson:01}. As a result, the sum of square differences cannot be used for this purpose optimally.

To tackle the aforementioned problem, we propose in this paper a multicomponent entropic similarity measure for DTI registration. A general framework for image registration methods relies on information-theoretic measures such as mutual information and Jensen-Shannon divergence~\cite{Hamza:03}. By employing the Jensen-Tsallis (JT) similarity measure~\cite{Khader:11,Khader:12}, the nonlinear intervoxel intensity differences are taken into account without the need for an explicit tensor reorientation during the optimization procedure. Hence, the tensors are only reoriented after the application of the final deformation field. More precisely, we propose a nonrigid image registration method by optimizing a multicomponent JT similarity measure using the quasi-Newton L-BFGS-B method~\cite{Nocedal:02} as an optimization scheme and cubic B-splines for modeling the nonrigid deformation field between the fixed and moving 3D image pairs. The analytical gradient of the multicomponent JT similarity is derived in a effort to design an efficient and accurate nonrigid registration algorithm. In order to achieve a compromise between the nonrigid registration accuracy and the associated computational cost, we implement a three-level hierarchical multi-resolution approach such that the image resolution is increased, along with the resolution of the control mesh, in a coarse to fine fashion. Since the JT is a robust measure of the image similarity, no tensor reorientation is performed in an iterative way. Tensor reorientation is only performed after the application of final deformation. A major advantage of not applying tensor reorientation iteratively is to decrease the computational complexity of the registration algorithm and hence the runtime. The experimental results demonstrate the registration accuracy of the proposed approach in comparison to the affine registration method based on mutual information~\cite{Leemans:05,Hecke:07}.

The rest of this paper is organized as follows. In Section 2, we provide a brief background on diffusion tensor imaging, followed by the problem formulation and the definition of the JT similarity measure. In Section 3, we describe in detail the proposed method, including the multicomponent JT similarity, and the tensor reorientation formulation. Then, we present a summary of our proposed algorithm. Section 4 provides experimental results on a diffusion tensor imaging data set to demonstrate the effectiveness and superior performance of our method compared to the affine registration technique.
\section{Background and Problem Formulation}

\subsection{Diffusion Tensor Imaging}
Water diffusion inside the brain can be characterized by a diffusion tensor, $\bm{D}$, at each voxel of an MRI volume. This diffusion tensor can be represented as a real, symmetric and positive definite matrix
\begin{equation}
\bm{D}=\left(
  \begin{array}{clcr}
               D_{xx} & D_{xy} & D_{xz}     \\
               D_{xy} & D_{yy} & D_{yz}    \\
               D_{xz} & D_{yz} & D_{zz}
          \end{array}
          \right).
\end{equation}
For each voxel, the signal intensity $S$ of the tissue is calculated as follows:
\begin{equation}
S=S_{0}\,e^{-b\,ADC}
\end{equation}
where $S_{0}$ is the signal intensity on the T2-weighted image, $b$ is a scalar weighting factor representing the strength of diffusion sensitivity, and $ADC$ is the apparent diffusion coefficient. $ADC$ is the projection of the diffusion tensor along the gradient of measure and describes the diffusivity along that particular direction. The diffusion tensor, $\bm{D}$, and the apparent diffusion coefficient, $ADC$ are related by the equation:
\begin{equation}
ADC = \hat{\bm{g}}_k^T\bm{D}\hat{\bm{g}}_{k},
\end{equation}
where $\hat{\bm{g}}_k$ is a dimensionless unit vector given by the direction of the measurement~\cite{Zhang:06}.

\noindent{\textbf{Acquisition and computation of the diffusion tensor:}} Several diffusion weighted (DW) images and different non-collinear gradient directions $\bm{g}_k (k=1,2, \ldots,N)$ should be acquired to compute the
diffusion tensor $\bm{D}(\bm{r})$, where $\bm{r}$ denotes the voxel position. Because $\bm{D}(\bm{r})$ is characterized by six degrees of freedom due to the symmetry of the tensor, at least six DW measurements $S_k(\bm{r})$ are needed, along with a reference image $S_0(\bm{r})$ acquired without diffusion weighting. In general, $\bm{D}(\bm{r})$ can be calculated for each voxel at position $\bm{r}$ by solving the following system of equations
\begin{equation}
S_k(\bm{r})=S_0(\bm{r})e^{-b\,\hat{\bm{g}}_k^T
\bm{D}(\bm{r})\hat{\bm{g}}_k} ~~\textrm{with}~~
\hat{\bm{g}}_k= \frac{\bm{g}_k}{\Vert \bm{g}_k \Vert}.
\label{eq:CompDT}
\end{equation}
Six axial DW measurements $S_{k}(\bm{r})$ and one non-DW image $S_{0}(\bm{r})$ are shown in Figure~\ref{fig:DWI}, along with the corresponding magnetic field gradients $\bm{g}_k\,(k = 1,\ldots, 6)$. Note the difference in intensity values for different gradient directions.

\begin{figure}[htbp]
\setlength{\tabcolsep}{.01em}
\centering
\begin{tabular}{cccc}
\includegraphics[width=1.5in,height=1.5in]{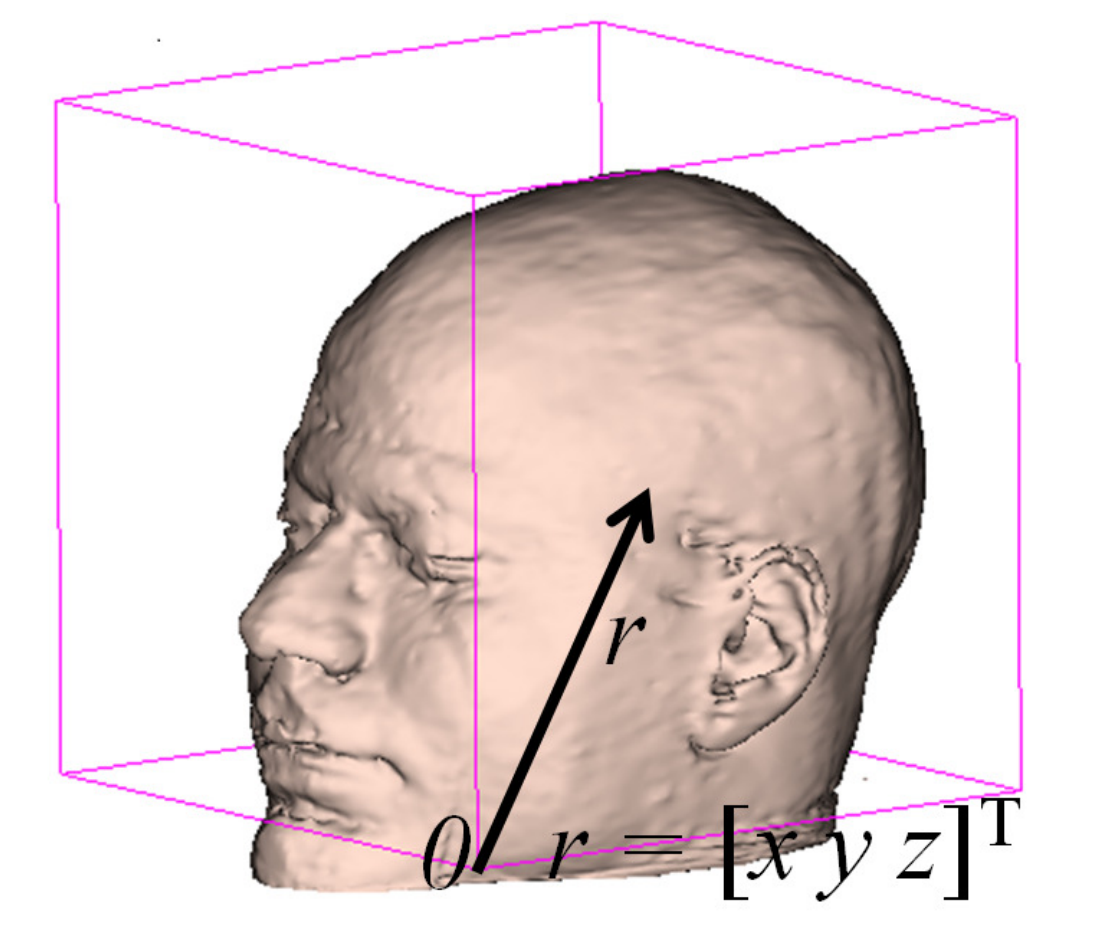}&
\includegraphics[width=1.3in,height=1.5in]{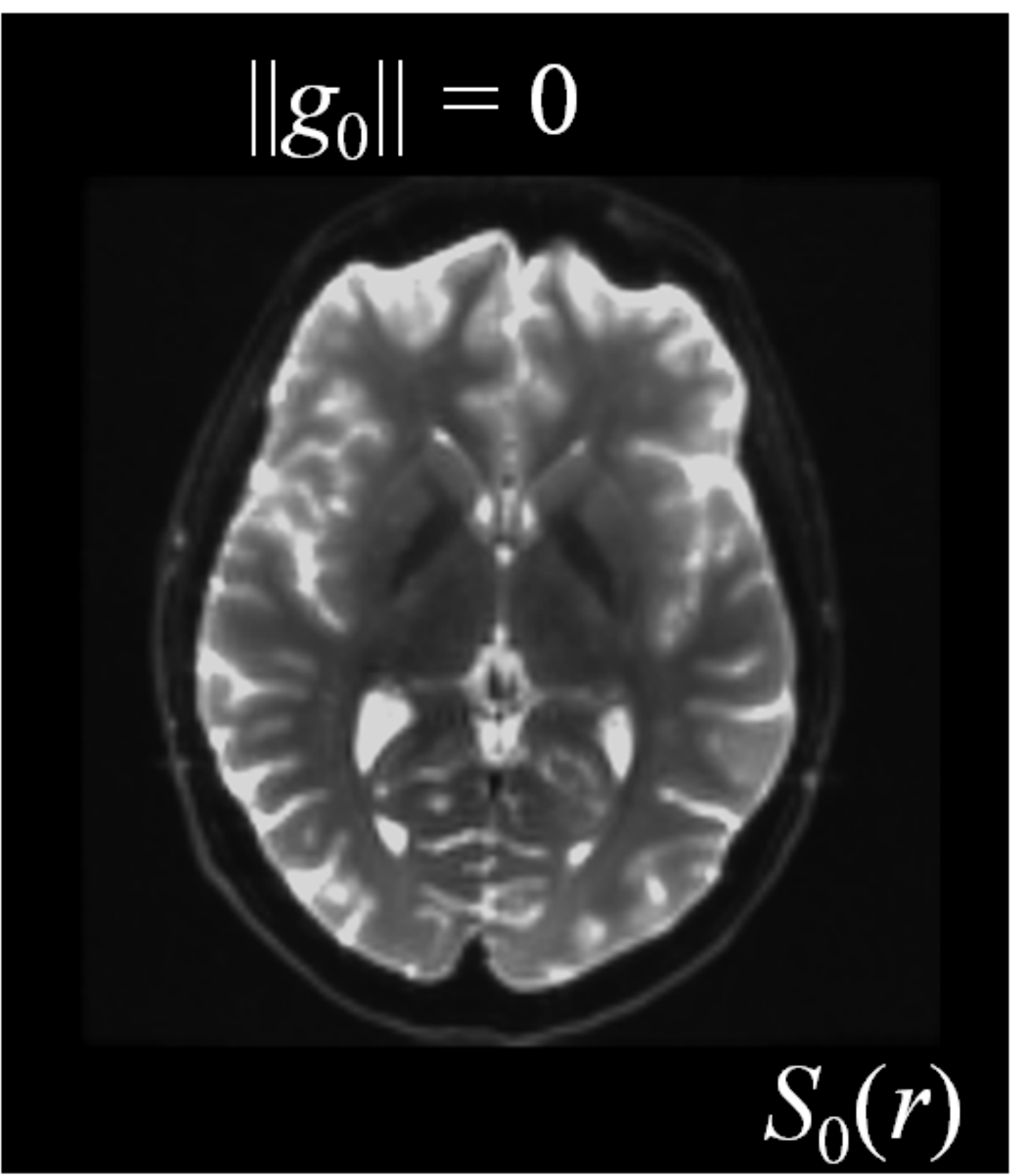}&
\includegraphics[width=1.3in,height=1.5in]{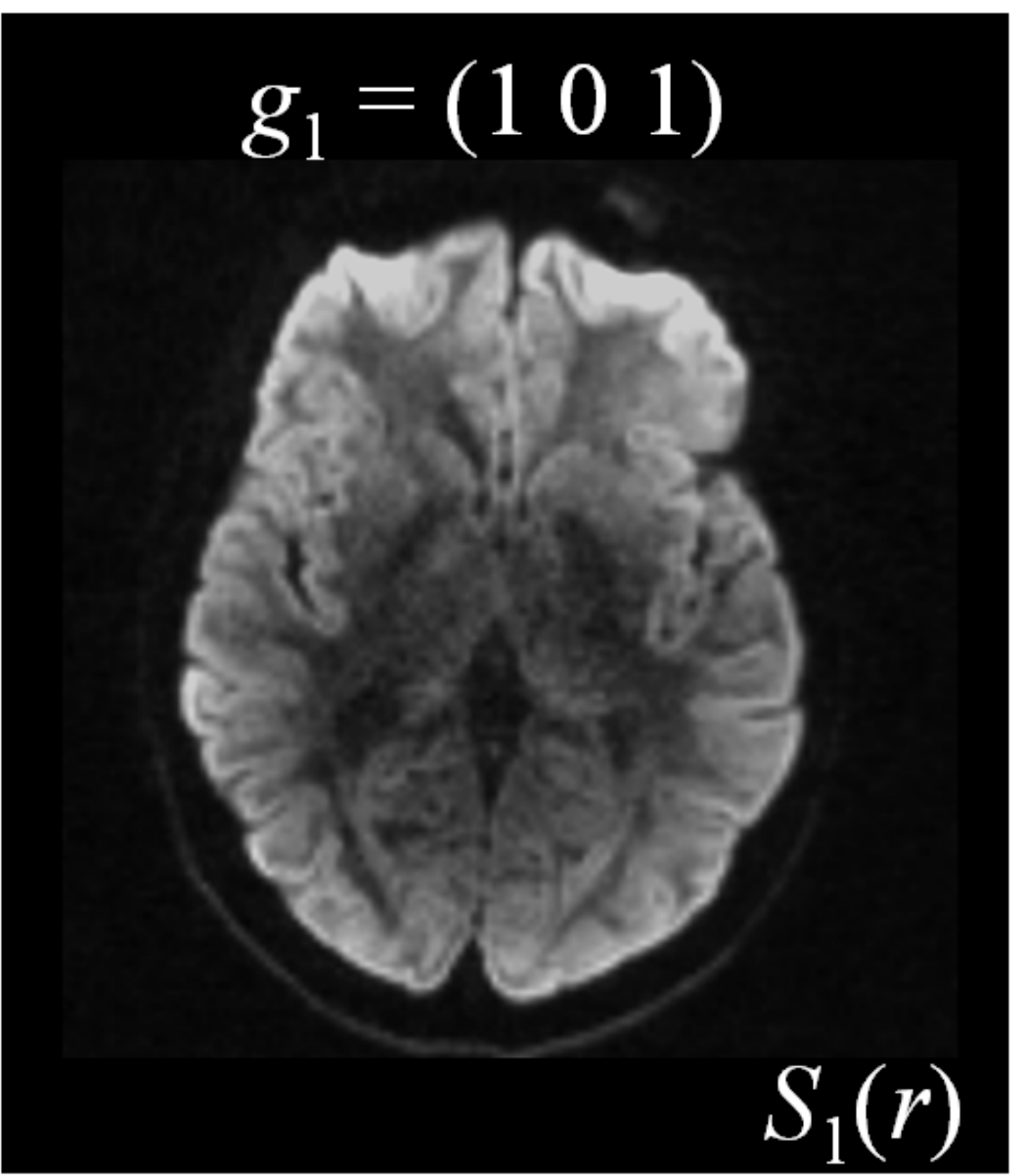}&
\includegraphics[width=1.3in,height=1.5in]{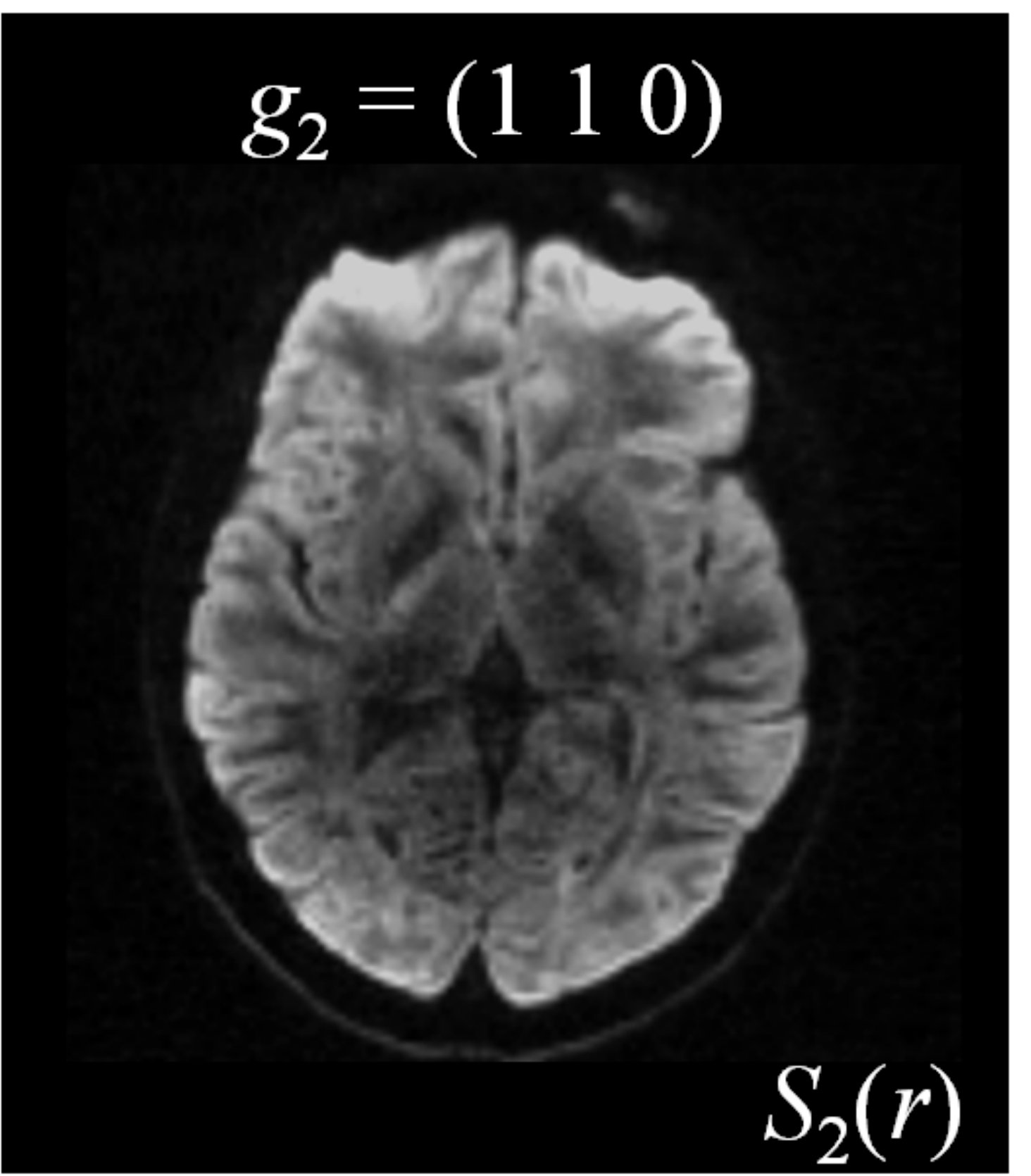}\\
(a) & (b) & (c) & (d) \\
\includegraphics[width=1.3in,height=1.5in]{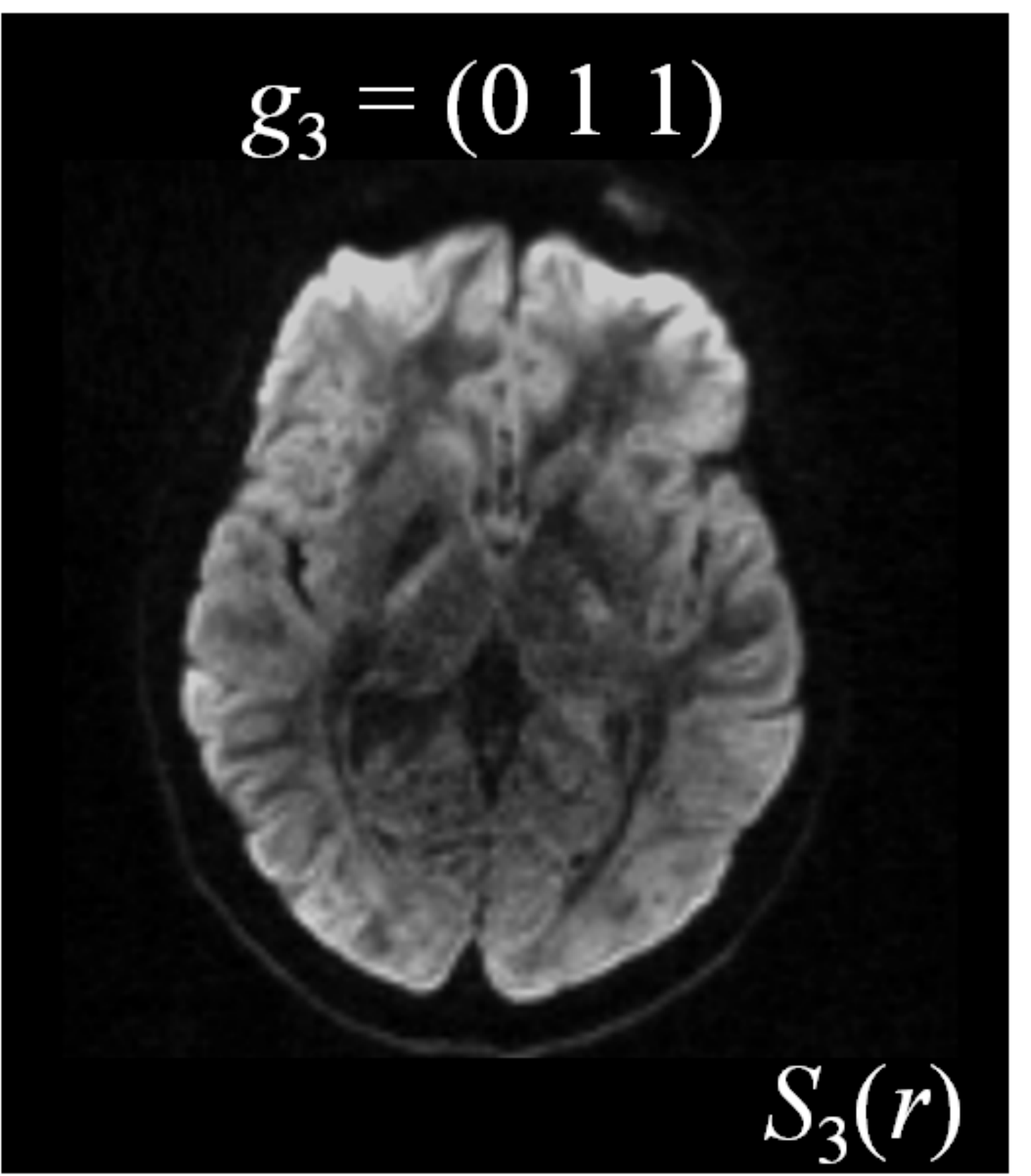}&
\includegraphics[width=1.3in,height=1.5in]{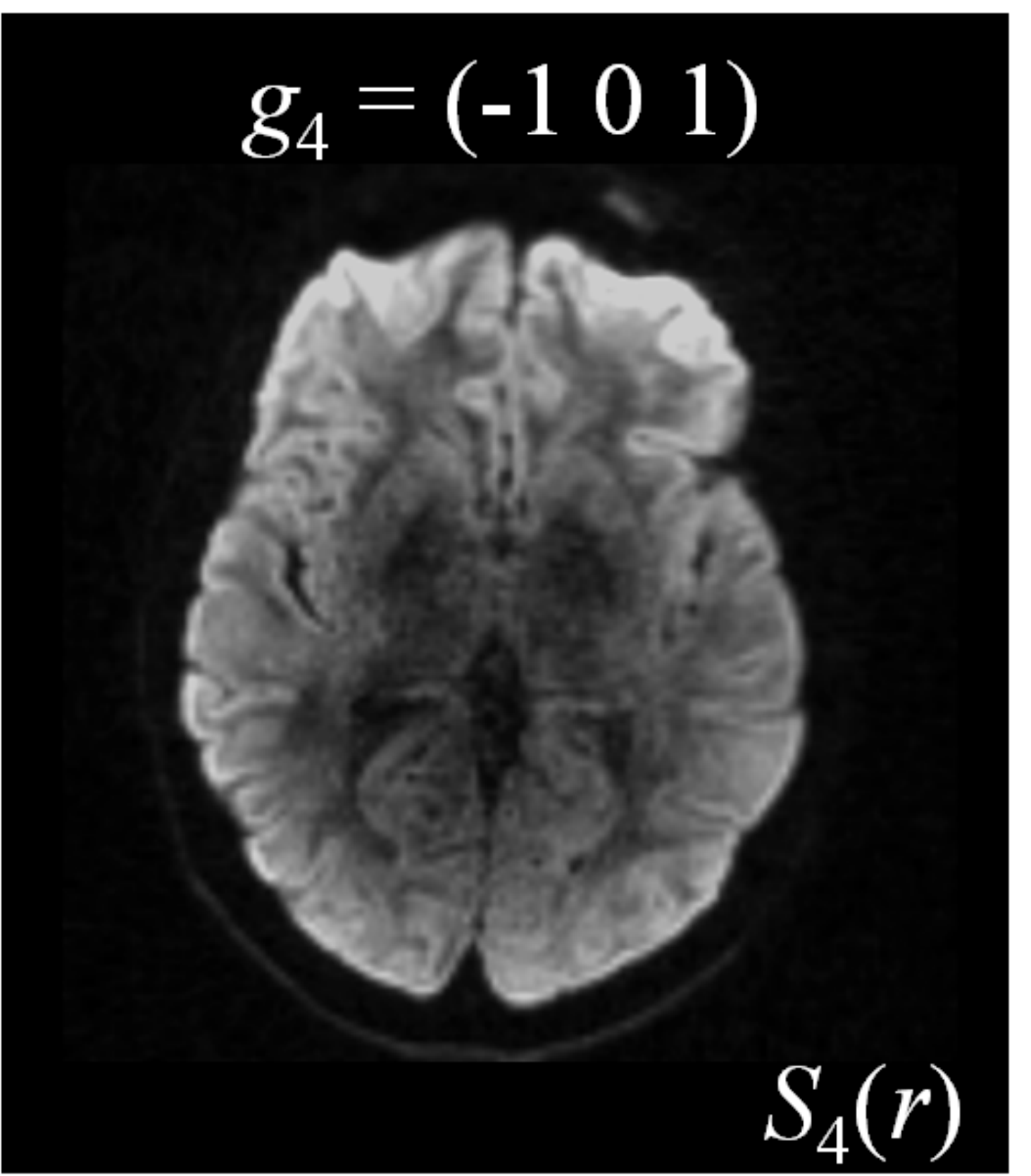}&
\includegraphics[width=1.3in,height=1.5in]{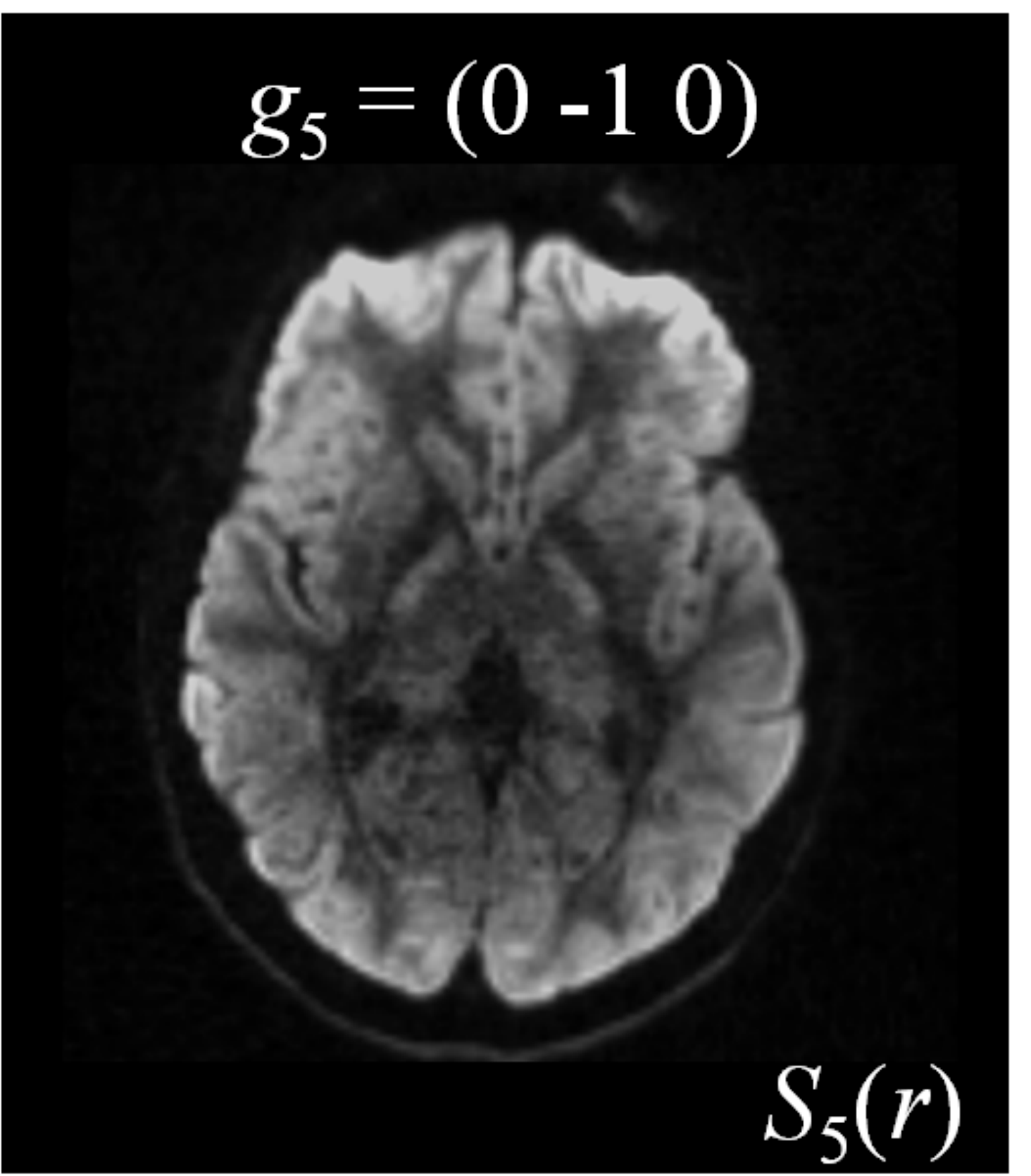}&
\includegraphics[width=1.3in,height=1.5in]{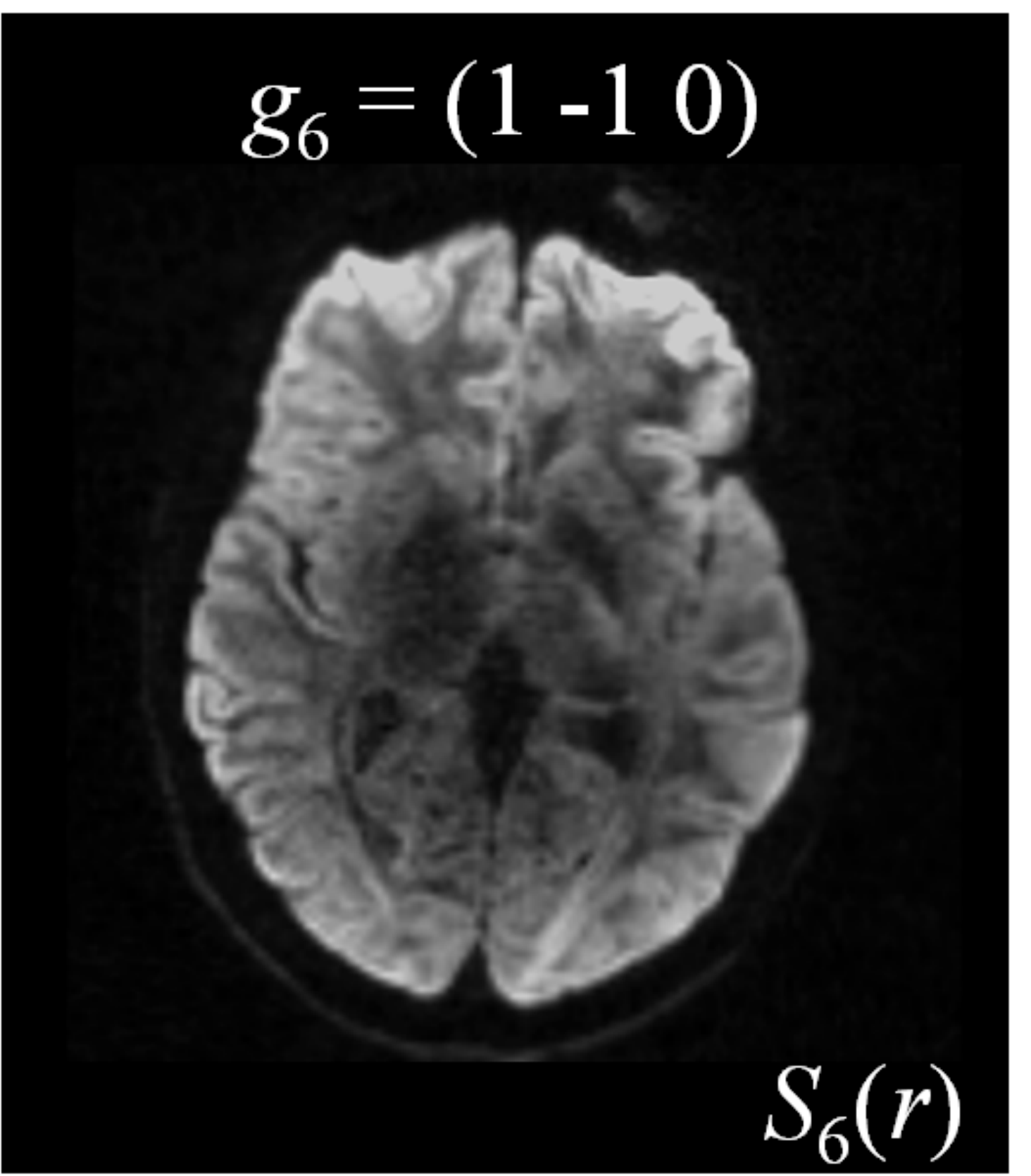}\\
(e) & (f) & (g) & (h)
\end{tabular}
\caption{Axial DW images $S_{k}(\bm{r})$ of the human brain for different gradient directions $\bm{g}_k$.}
\label{fig:DWI}
\end{figure}
The diffusion tensor field $\bm{D}$ is in fact a covariance matrix describing the translational displacement of the diffusing molecules. Therefore, an ellipsoidal shape can be associated with $\bm{D}$, which represents the probabilistic iso-surface of this molecular diffusion~\cite{Taylor:04}. Because $\bm{D}$ is a symmetric and positive definite second-order tensor, its spectral decomposition may be written as
\begin{equation}
\bm{D}=\lambda_{1}\bm{e}_{1}\bm{e}_{1}^{T}+\lambda_{2}\bm{e}_{2}\bm{e}_{2}^{T}
+\lambda_{3}\bm{e}_{3}\bm{e}_{3}^{T}
\end{equation}
where $\lambda_1 \ge\lambda_2 \ge \lambda_3 >0$ are the positive eigenvalues of $\bm{D}$ and $\bm{e}_i$ are the associated orthonormal eigenvectors. These eigenvectors and eigenvalues represent the principal axes of the ellipsoid and their corresponding principal diffusion coefficients, respectively, as illustrated in Figure~\ref{Fig:diffusioncases}. Therefore, the ellipsoid axes are oriented according to the tensor eigenvectors, and their lengths depend on the tensor eigenvalues.
\begin{figure}[htb]
\centerline{
\begin{tabular}{ccc}
\includegraphics[width=1.5in,height=1.5in]{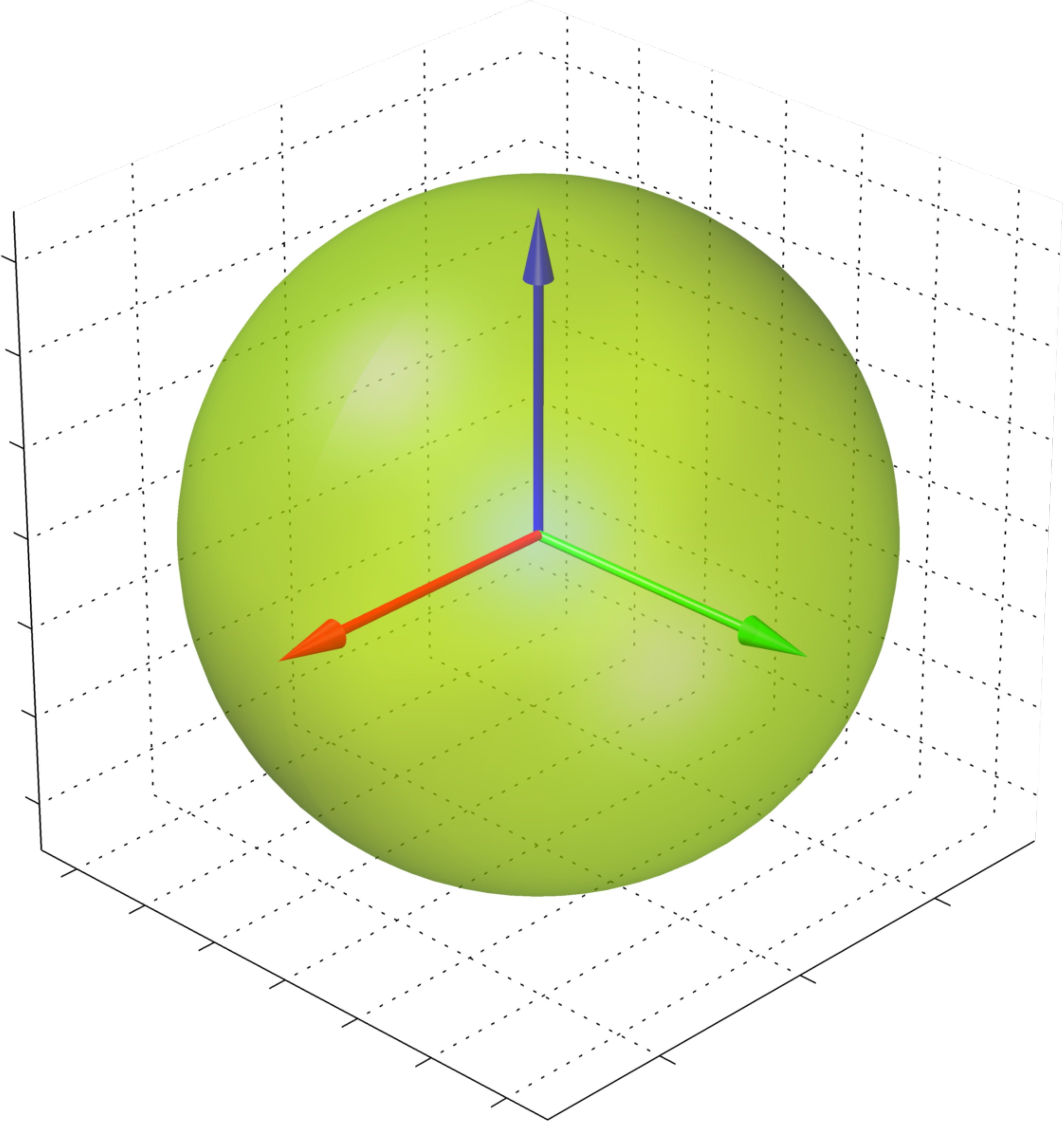} &
\includegraphics[width=1.5in,height=1.5in]{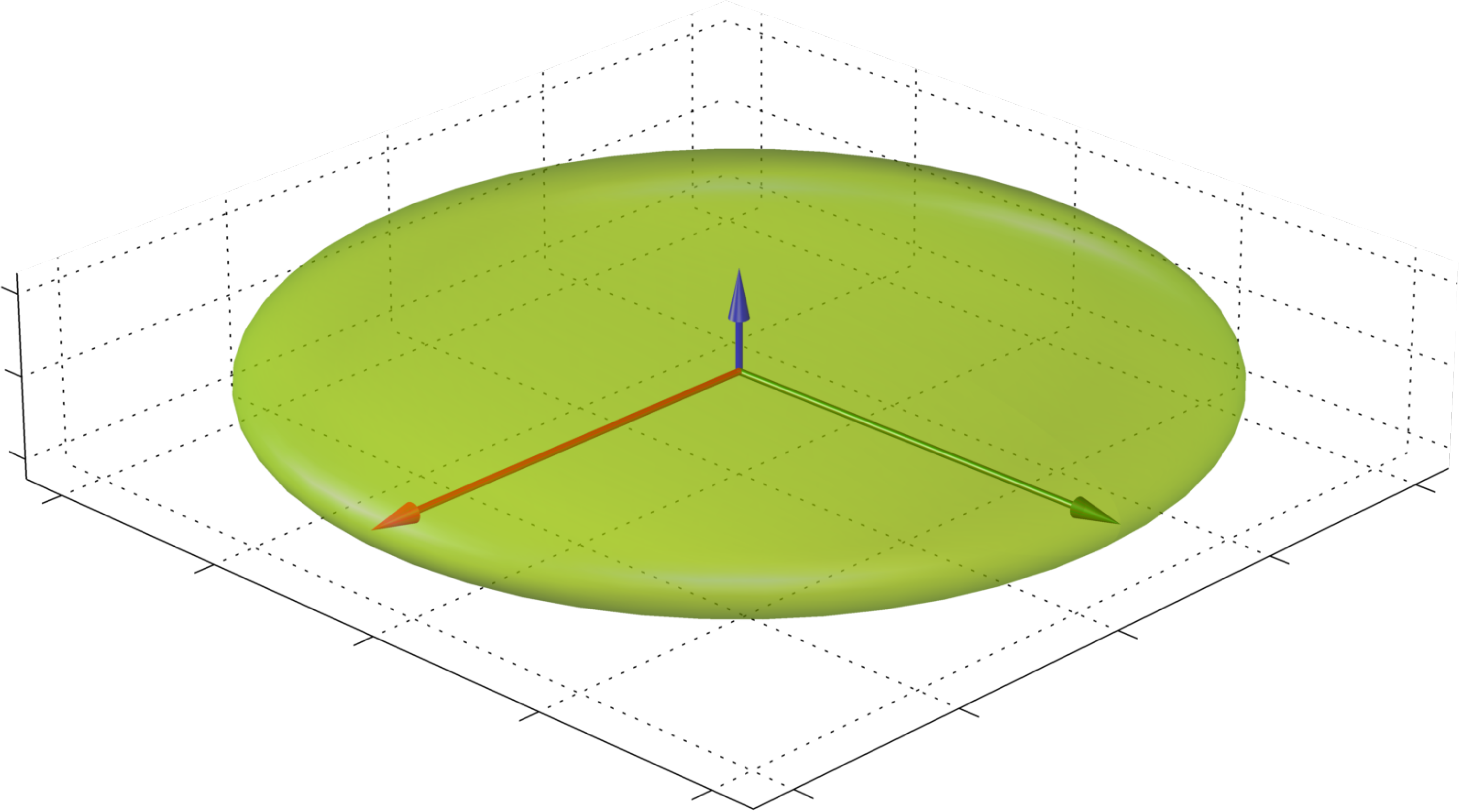}&
\includegraphics[width=1.5in,height=1.5in]{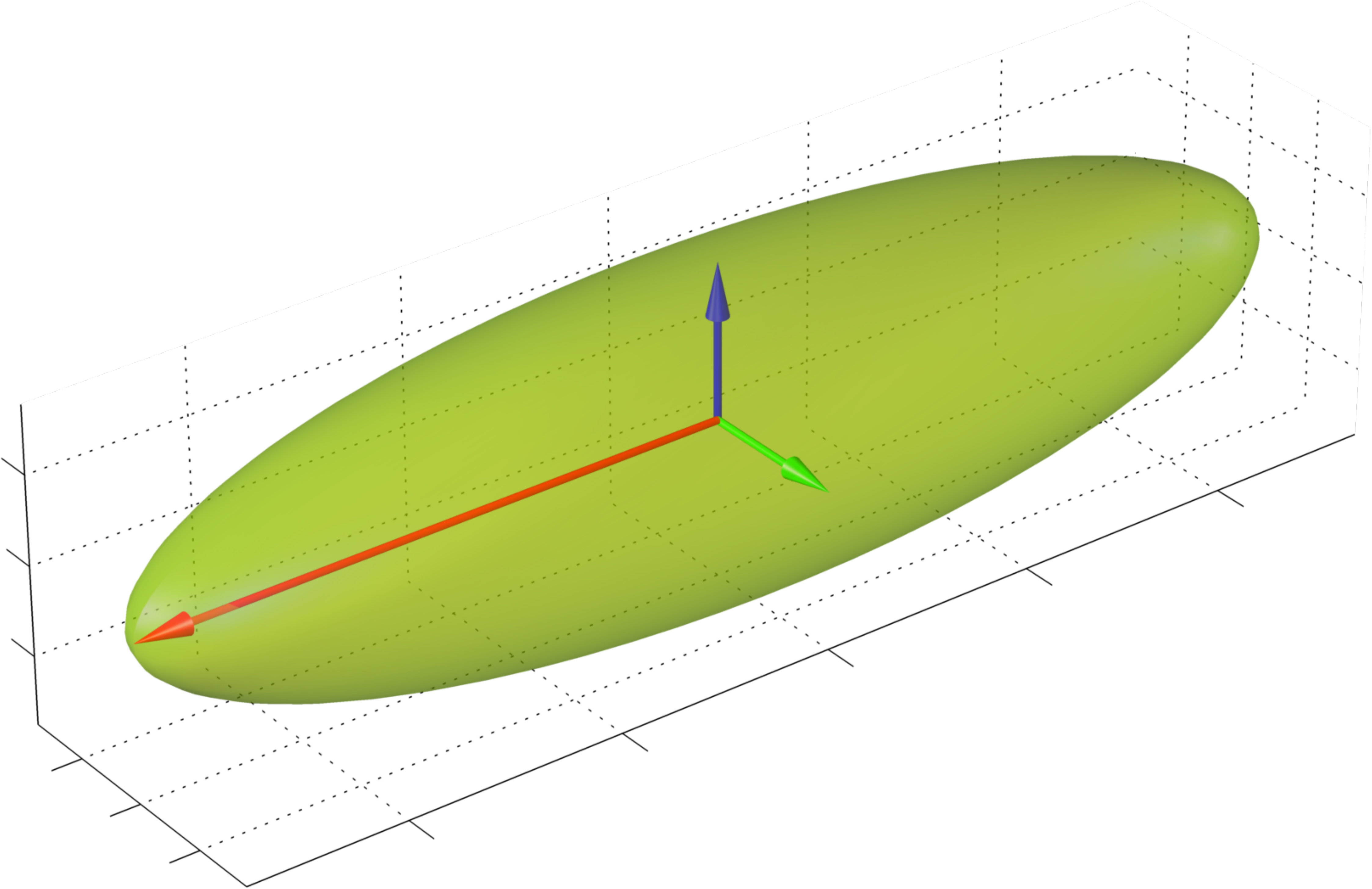} \\
$\lambda_{1}\approx\lambda_{2}\approx\lambda_{3}$ & $\lambda_{1}\approx\lambda_{2}\gg\lambda_{3}$ &
$\lambda_{1}\gg\lambda_{2}\approx\lambda_{3}$
\end{tabular}}
\caption{Different cases of diffusion: Spherical diffusion (left); planar diffusion (center); linear diffusion (right).}
\label{Fig:diffusioncases}
\end{figure}
\subsection{Scalar Indices}
The first eigenvector $\bm{e}_1=(e_{1x},e_{1y},e_{1z})$, also called principal diffusion vector, of $\bm{D}$ describes the predominant diffusion direction, which is parallel to the orientation of the corresponding underlying WM fiber system. Figure~\ref{fig:DTIFull} shows a visualization of the color-coded MR-DTI data with ellipsoids. The predominant diffusion direction can be directly related to a Green (G), Red (R) and Blue (B) digital color triple. The convention in which the G, R and B color components represent the directions is as follows:
\begin{equation}
 \left[\lVert\bm{e}_{1x}\rVert, \lVert\bm{e}_{1y}\rVert, \lVert\bm{e}_{1z}\rVert\right] = \left[\text{G}, \text{R}, \text{B}\right].
 \label{eq:DTdecomp}
\end{equation}
The RGB color-coded directionality maps provide an indication of the direction in which water diffusion is the highest and improve the visibility of different WM fiber bundles.

\noindent{\textbf{Trace and Mean Diffusivity:}} The total diffusivity is $\textrm{trace}(\bm{D}) = \sum^{3}_{i=1}\lambda_{i}$, and the mean diffusivity (MD) is equal to one third of $\textrm{trace}(\bm{D})$. The MD measure serves as an indicator of brain maturation and/or injury, and provides the overall magnitude of water diffusion independent of anisotropy~\cite{Mori:95}. The MD map is shown in Figure~\ref{fig:test1}(c), where higher values of average diffusion appear brighter.

\noindent{\textbf{Fractional Anisotropy (FA):}} FA serves as an indicator of the degree of water diffusion anisotropy independent of the overall water diffusion coefficient and is defined as
\begin{equation}
\mathrm{FA}=\sqrt{\frac{3}{2}\frac{\sum^{3}_{i=1}
{(\lambda_{i}-\frac{1}{3}\textrm{trace}(\bm{D}))^2}}{\sum^{3}_{i=1}{\lambda_{i}^2}}},
\end{equation}
which is basically the normalized standard deviation of the eigenvalues. The values of FA vary from 0 to 1 with higher values corresponding to greater diffusion anisotropy. Figure~\ref{fig:test1}(c) shows the FA map of the same slice as in Figure~\ref{fig:test1}(a). The higher values of FA correspond to the WM regions containing densely packed fiber bundles that cause anisotropic diffusion by restricting water movement along the direction perpendicular to the fiber bundles.

\begin{figure}[htbp]
\centering
\includegraphics[scale=.6]{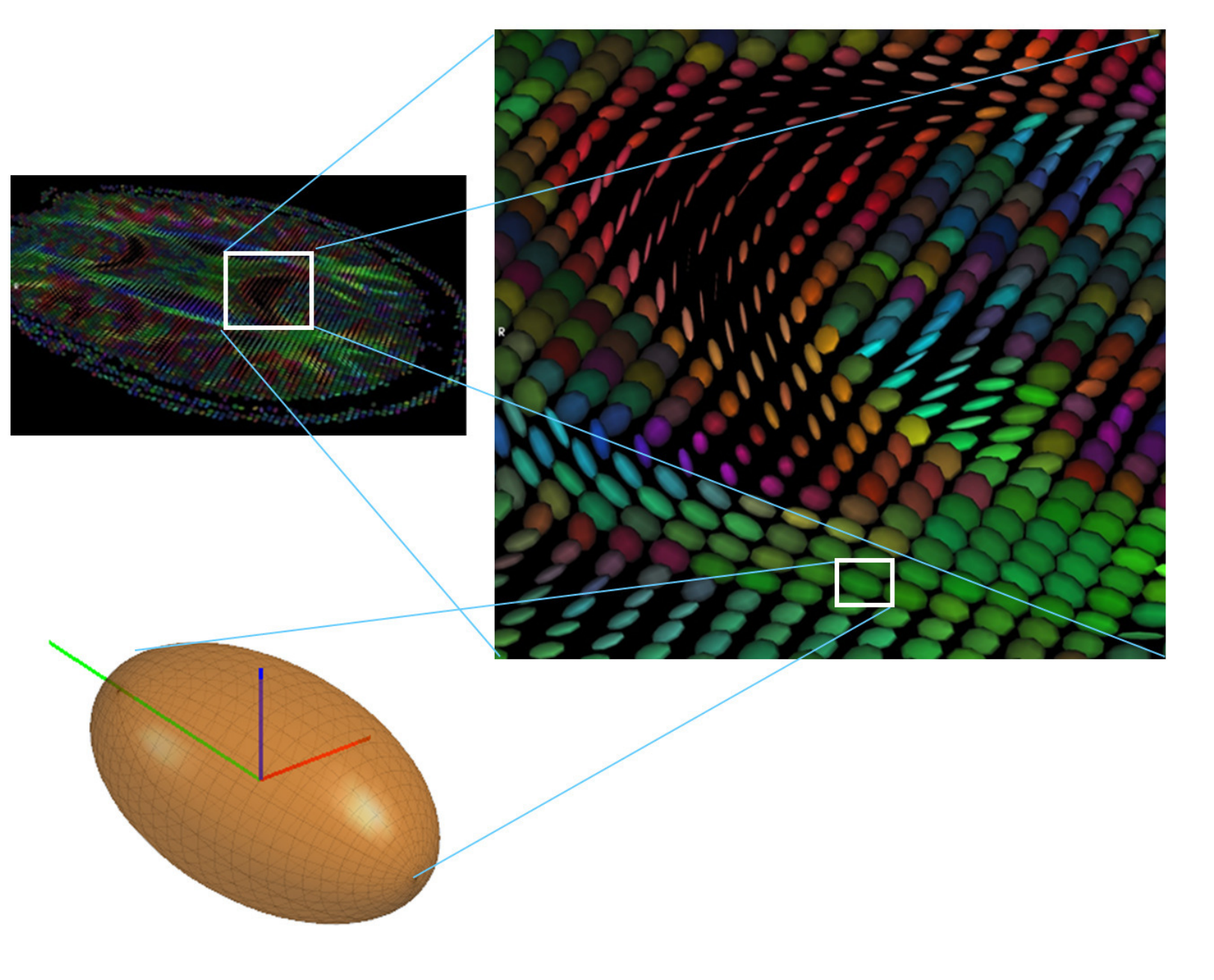}
\caption{Ellipsoidal representation of the diffusion tensor at each voxel location of a DTI image.}
\label{fig:DTIFull}
\end{figure}

\begin{figure}[htbp]
\setlength{\tabcolsep}{.01em}
\centering
\begin{tabular}{cccc}
\includegraphics[width=1.3in,height=1.5in]{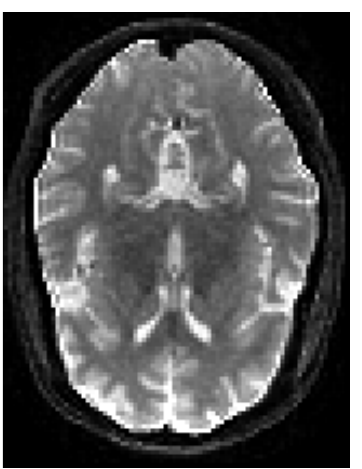}&
\includegraphics[width=1.3in,height=1.5in]{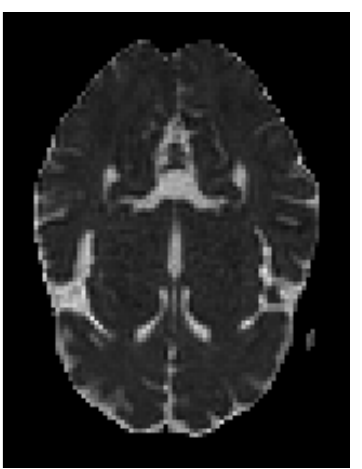}&
\includegraphics[width=1.3in,height=1.5in]{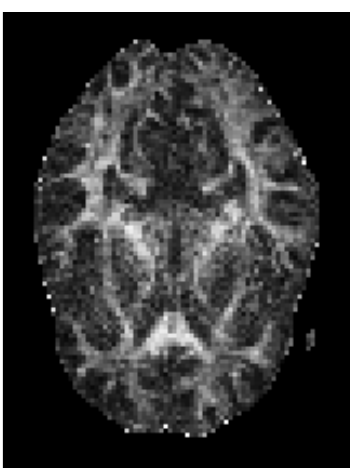}&
\includegraphics[width=1.3in,height=1.5in]{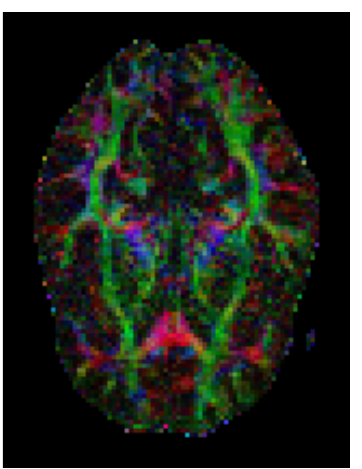}\\
(a) & (b) & (c) & (d)
\end{tabular}
\caption{Axial slice No. 30 of the DT image chosen as the template in this study : (a) DW image; (b) mean diffusion; (c) fractional anisotropy; (d) color-coded DT elements. The DT maps are color-coded according to the diffusion direction.}
\label{fig:test1}
\end{figure}

\subsection{Problem Statement}
Let $I$ and $J$ be two misaligned images to be registered, where $I$ is the fixed image and $J$ is the moving image. The moving image $J$ is obtained by applying a deformation field $\bm{\Phi}$ to the fixed image $I$, as depicted in Figure~\ref{fig:problem4}. Note that the deformation field $\bm{\Phi}$ can be applied directly to the DT components of the fixed image $I$, or to the DW images before calculating the DT components. The deformation field $\bm{\Phi}$ is described by a transformation function $g(\bm{x};\bg{\mu}): V_J \to V_I$, where $V_J$ and $V_I$ are continuous domains on which $J$ and $I$ are defined, and $\bg{\mu}$ is a vector of transformation parameters to be determined.

The image alignment or registration problem may be formulated as an optimization problem:
\begin{equation}
\hat{\bg{\mu}} = \arg\min_{\bg{\mu}} \mathcal{S}\bigl(I(\bm{x}),J(g(\bm{x};\bg{\mu}))\bigr),
\label{eq:SimOptimization}
\end{equation}
where $\mathcal{S}(\cdot,\cdot)$ is a cost function that measures the similarity between the fixed image and the deformed moving image.

To align the transformed moving image $J(g(\bm{x};\bg{\mu}))$ to the fixed image $I$, we seek the vector of transformation parameters $\bg{\mu}$ that minimize the cost function $\mathcal{S}\bigl(I(\bm{x}),J(g(\bm{x};\bg{\mu}))\bigr)$.
\begin{figure}[htbp]
\setlength{\tabcolsep}{.01em}
\centering
\begin{tabular}{ccc}
\includegraphics[width=1.3in,height=1.5in]{refDTI_ORIGINAL_SIZE.pdf}&
\includegraphics[width=1.3in,height=1.5in]{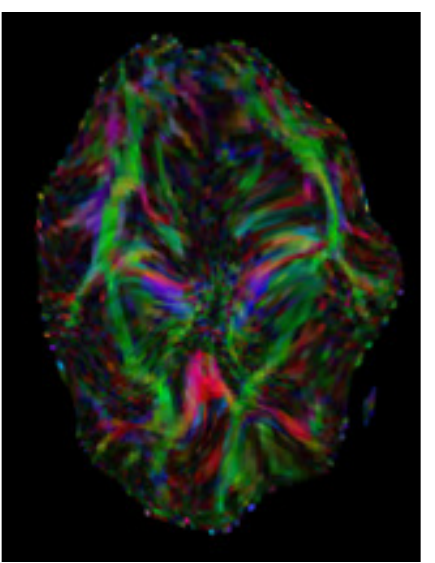}&
\includegraphics[width=1.3in,height=1.5in]{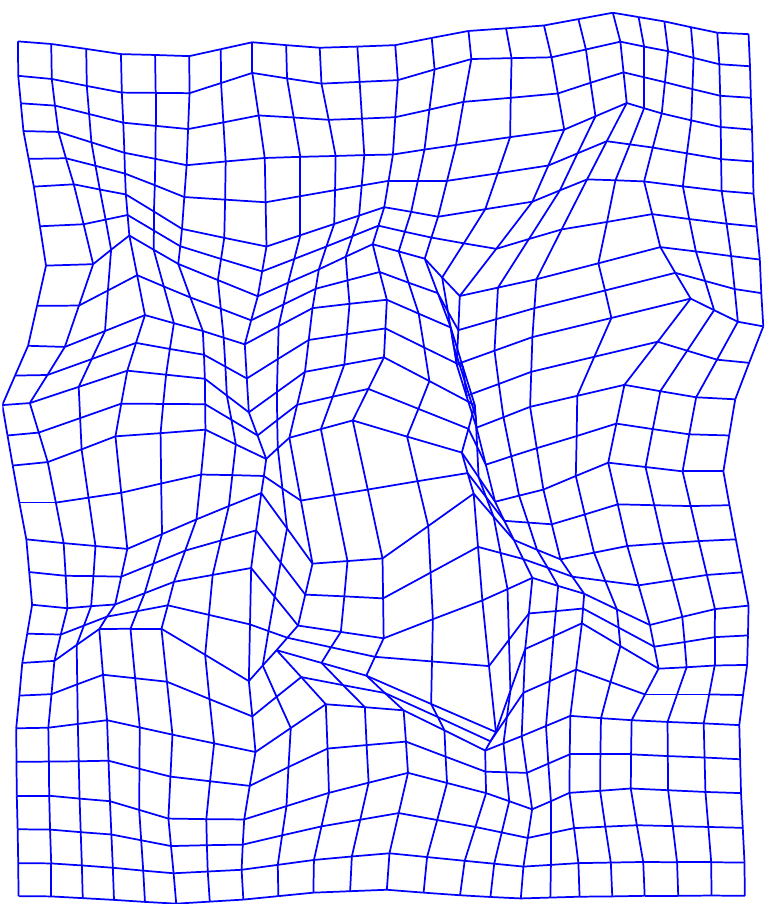}\\
(a) & (b) & (c)
\end{tabular}
\caption{(a) Fixed image $I$; (b) moving image $J$; (c) deformation field $\bm{\Phi}$.}
\label{fig:problem4}
\end{figure}
\subsection{Jensen-Tsallis Similarity Measure}
Recently, there has been a concerted research effort in statistical physics to explore the properties of Tsallis entropy, leading to a statistical mechanics that satisfies many of the properties of the standard
theory~\cite{Tsallis:88}. Tsallis entropy is defined as
\begin{equation}
\label{tsallis}
H_{\alpha}(\bm{p})=
\frac{1}{1-\alpha}\Bigl(\sum_{j=1}^{k}p_{j}^{\alpha}-1\Bigr) =
-\sum_{j=1}^{k}p_{j}^{\alpha}\log_{\alpha}(p_{j}),
\end{equation}
where $\bm{p}=(p_1,p_2,\ldots,p_k)$ is a probability distribution, $\log_{\alpha}$ is the $\alpha$-logarithm function defined as $\log_{\alpha}(x)=(1-\alpha)^{-1}(x^{1-\alpha}-1)$ for $x>0$, and $\alpha\in (0,1)\cup (1,\infty)$ is an exponential order (also referred to as entropic index).

The Jensen-Tsallis (JT) similarity measure~\cite{Khader:11} between $n$ probability distributions $\bm{p}_1,\bm{p}_2,\ldots,\bm{p}_n$ is given by
\begin{equation}
\mathcal{S}_{\alpha}^{\bg{\omega}}(\bm{p}_1,\ldots,\bm{p}_n)
=1-\frac{\mathcal{D}_{\alpha}^{\bg{\omega}}(\bm{p}_1,\ldots,\bm{p}_n)}{\log_{\alpha}n},
\label{eq:similarityJensenTsallis}
\end{equation}
where $\mathcal{D}_{\alpha}^{\bg{\omega}}$ is the JT divergence defined as
\begin{equation}
\label{eq:Jensen-TsallisD}
\mathcal{D}_{\alpha}^{\bg{\omega}}(\bm{p}_1,\ldots,\bm{p}_n)=
H_{\alpha}\left(\sum_{i=1}^{n}\omega_{i}\bm{p}_{i}\right)
-\sum_{i=1}^{n}\omega_{i}H_{\alpha}(\bm{p}_{i})
\end{equation}
and $\bg{\omega}=(\omega_1,\omega_2,\ldots,\omega_n)$ is a nonnegative weight vector such that $\sum_{i=1}^{n}\omega_{i}=1$.

Figure~\ref{fig:JTsimilarity} illustrates the JT similarity between two Bernoulli distributions $\bm{p}=(p,1-p)$ and $\bm{q}=(1-p,p)$, with uniform weight $\omega_1=\omega_2=1/2$, for different values of the entropic index.
\begin{figure}
\centering
\includegraphics[scale=.7]{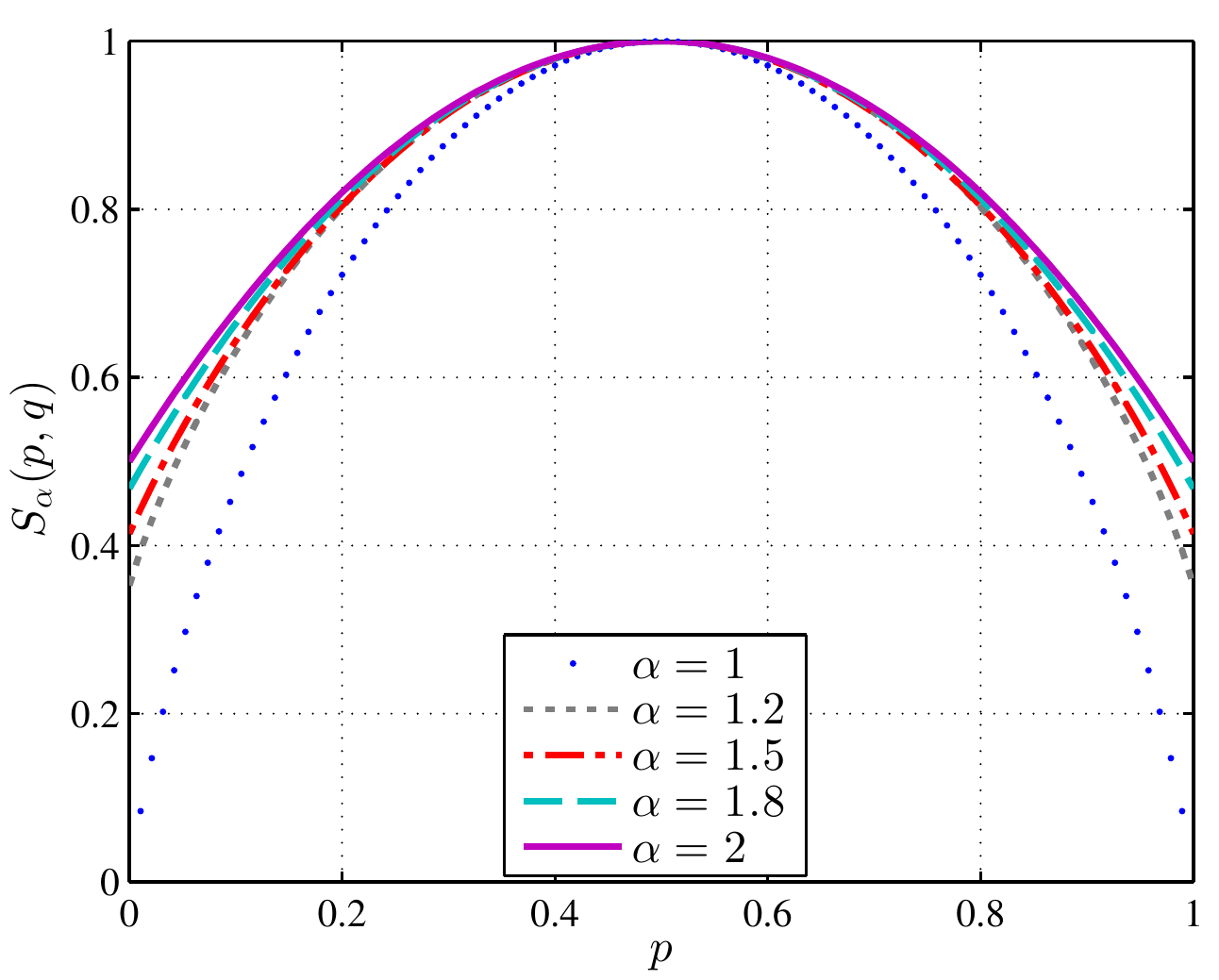}
\caption{JT similarity $S_{\alpha}(\bm{p},\bm{q})$ between two Bernoulli distributions $\bm{p}=(p,1-p)$ and $\bm{q}=(1-p,p)$ for different values of $\alpha$.}
\label{fig:JTsimilarity}
\end{figure}
As can be seen in Figure~\ref{fig:JTsimilarity}, the highest similarity corresponds to the entropic index $\alpha=2$. Consequently, we choose an entropic index $\alpha=2$ throughout the paper unless indicated otherwise.
\section{Proposed Framework}
We propose two different registration approaches using a different number of components, namely the FA map with one component, and the DT elements with six components. In the first approach, referred to as JT-FA, we apply our JT similarity based registration algorithm to the FA map between the fixed and deformed moving images. In this approach we do not need to modify the JT similarity measure because each DTI data set contains only one FA map (i.e. $L=1$).

In the second approach, which we call JT-DT, the JT similarity measure needs to be modified to register the multicomponent DT elements because each DTI data set contains six DT elements $L=6$ (i.e. $\ell=1,\ldots,L$). To determine the multicomponent JT similarity, the JT measure is computed for all corresponding components separately, assuming that they are independent. For instance, the first DT element image (i.e. for $\ell=1$) of the fixed DTI data set is compared to the first DT element image of the moving DTI data set. A similar approach was presented in~\cite{Hecke:07} using mutual information as a similarity metric. The multicomponent JT similarity measure is then calculated by averaging the JT similarity of the different corresponding components, as explained in the next subsection. Finally, the JT-DT approach is optimized via an iterative process.
\subsection{Multicomponent Jensen-Tsallis Similarity}
As mentioned earlier, we assume that the entropic index is set to $\alpha=2$. For each component $\ell=1,\ldots,L$, the JT similarity, denoted by $\mathcal{S}_{\ell,\alpha}^{\bg{\omega}}$, and its derivative between all corresponding components of the moving and fixed DTI data sets are given by
\begin{equation}
\label{eq:jensen-tsallisDTI}
     \mathcal{S}_{\ell,2}^{\bg{\omega}}(\bm{p}_{\ell,1},\ldots,\bm{p}_{\ell,n})=\displaystyle
     1-\frac{\mathcal{D}_{\ell,2}^{\bg{\omega}}(\bm{p}_{\ell,1},\ldots,\bm{p}_{\ell,n})}{\log_{2}n}
\end{equation}
and
$$
\displaystyle\frac{\partial
\mathcal{S}_{\ell,2}^{\bg{\omega}}(\bm{p}_{\ell,1},\ldots,\bm{p}_{\ell,n})}{\partial\bm{\mu_\ell}}=\displaystyle
       -\frac{\partial \mathcal{D}_{\ell,2}^{\bg{\omega}}(\bm{p}_{\ell,1},\ldots,\bm{p}_{\ell,n})}{\partial\bm{\mu_\ell}} \frac{1}{\log_{2}n}
$$
respectively, where
$$\bm{p}_{\ell,i} = \bm{p}_{\ell,i}\bigl(J(g(\bm{x};\bg{\mu}))|I(\bm{x})\bigr),\quad \forall i=1,\ldots, n,$$
are the conditional intensity probability distributions of the corresponding $\ell$-th image component.

The multicomponent JT similarity measure is obtained by averaging the JT similarity for multiple components. In other words, the multicomponent JT similarity, denoted by $\mathcal{S}_{L,2}^{\bg{\omega}}$, and its derivative are given by
\begin{equation}
\label{eq:jensen-tsallisDTI-AVG}
     \mathcal{S}_{L,2}^{\bg{\omega}}(\bm{p}_1,\ldots,\bm{p}_n)=
     \displaystyle\frac{1}{L}\sum_{\ell=1}^{L}\mathcal{S}_{\ell,2}^{\bg{\omega}}(\bm{p}_{\ell,1},\ldots,\bm{p}_{\ell,n})
\end{equation}
and
\begin{equation}
\label{eq:jensen-tsallisDTI-grad}
\displaystyle\frac{\partial
\mathcal{S}_{L,2}^{\bg{\omega}}(\bm{p}_1,\ldots,\bm{p}_n)}{\partial\bg{\mu}}=
\displaystyle\frac{1}{L}\sum_{\ell=1}^{L}\frac{\partial
\mathcal{S}_{\ell,2}^{\bg{\omega}}(\bm{p}_{\ell,1},\ldots,\bm{p}_{\ell,n})}{\partial\bm{\mu_\ell}}
\end{equation}
Note that when $L=1$, the multicomponent JT similarity reduces to the JT measure.

To solve the nonrigid DT image alignment problem given by Eq.~(\ref{eq:SimOptimization}), we will use the multicomponent JT similarity measure as a matching criterion. According to Eq.~(\ref{eq:SimOptimization}), if $I$ and $J$ are scalar-valued images, then image transformations only change the position of each voxel $\bm{x}$. Image deformation is more complex for diffusion tensor images because the transformations also change the diffusion tensor orientation. Hence, tensor reorientation is needed to ensure that DT orientation is consistent with the underlying deformed microstructure.
\subsection{Tensor Reorientation Formulation}
For rigid transformation of DT images, tensor reorientation is straightforward. Let the orthogonal matrix $\bm{R}$ denote the rotational component of the rigid transformation to each tensor. Thus, the reorientation on a diffusion tensor $\bm{D}$ is $\bm{D}'=\bm{R}\bm{D}\bm{R}^{T}$. On the other hand, for nonrigid transformations of DT images, when the moving image $J$ is deformed to match the fixed image $I$ with the mapping $g: V_{J} \rightarrow V_{I}$, the tensor at voxel location $\bm{x}$ is deformed according to the Jacobian matrix $\bm{M}=\nabla g^{-1}(\bm{x})$. Alexander {\em et al.}~\cite{Alexander:01} proposed a simple reorientation strategy, called finite strain method, to determine a rotational matrix $\bm{R}$ from the Jacobian matrix $M$. The finite strain algorithm selects the best orthogonal approximation of $\bm{M}$ to be $\bm{R}$, where $\bm{R}$ is the solution of $\arg\min_{\bm{R}'}||\bm{R}'-\bm{M}||$. Figure~\ref{fig:TR-Example} shows a registered image before and after applying FS tensor reorientation algorithm. In Figure~\ref{fig:TR-Example}(b), we display a portion of the diffusion tensor field before applying the tensor reorientation algorithm. Some tensor orientations in this field are not consistent with the anatomy after image deformation. Figure~\ref{fig:TR-Example}(d) shows that the orientations of all tensors become consistent with the anatomy after applying the finite strain method.

\begin{figure}[htb]
\centering
\includegraphics[scale=.4]{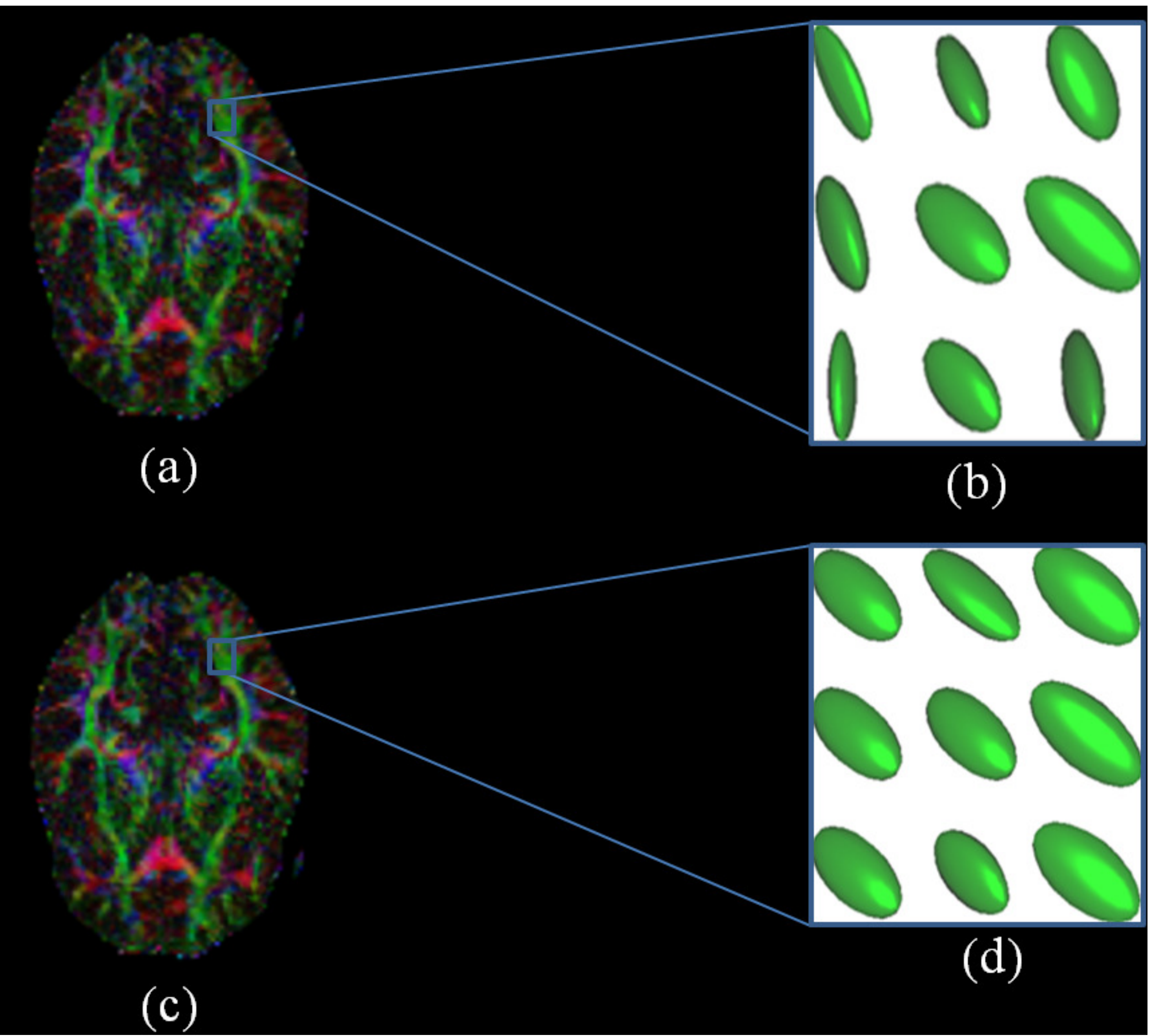}
\caption{(a) and (c) Registered images before and after tensor reorientation, respectively; (b) and (d) Diffusion tensors of a certain region before and after applying tensor reorientation, respectively.} \label{fig:TR-Example}
\end{figure}

\subsection{Transformation Model}
We model the transformation $g(\bm{x};\bg{\mu})$ using the free form deformation~\cite{Mattes:03,Khader:11}, which is based on cubic B-splines. Let $\bm{\Phi}$ denote a $n_x\times{n_y}\times{n_z}$ mesh of control points $\varphi_{i,j,k}$ with a uniform spacing $\Delta$. Then, the 3D transformation at any point $\bm{x}=[x, y, z]^T$ in the moving image is interpolated using a linear combination of cubic B-spline convolution kernels as follows
\begin{equation}
\label{splinetrans}
g(\bm{x};\bg{\mu})=\sum_{ijk}\eta_{ijk}\beta^{(3)}\left(\frac{\bm{x}-\varphi_{ijk}}{\Delta}\right),
\end{equation}
where $\beta^{(3)}(\bm{x})=\beta^{(3)}(x)\beta^{(3)}(y)\beta^{(3)}(z)$ is a separable cubic B-spline convolution kernel~\cite{Khader:11}, and $\eta_{ijk}$ are the deformation coefficients associated to the control points $\varphi_{ijk}$. The degree of nonrigidity can be adopted to a specific registration problem by varying the mesh spacing or the resolution of the mesh $\bm{\Phi}$ of control points. The parameter vector $\bg{\mu}=(\eta_{ijk})$ represents the vector of deformation coefficients associated to the control points $\varphi_{ijk}$, where the indices $i, j, k$ denote the coordinates of the control points on the mesh grid.

\subsection{Implementation}
The proposed algorithm for nonrigid DTI registration is implemented by changing the deformation in the moving image(s) until the discrepancy between the moving and fixed images is minimized. The main algorithmic steps of our DTI registration framework are summarized in Algorithm~1. First, the algorithm initializes the deformation field $\bm{\Phi}$ by creating a uniform B-spline control grid with predefined spacing knots. Next, a 3-level hierarchical multi-resolution scheme is used to achieve the best compromise between the registration accuracy and the associated computational cost. As the hierarchical level increases the resolution of the control mesh is increased, along with the image resolution, in a coarse to fine fashion. In each hierarchical level, a limited-memory, quasi-Newton minimization scheme is used to find the optimum set of transformation parameters that reduce the multicomponent JT cost function until the difference between the cost function values in two consecutive iterations is less than $\varepsilon = 0.01$. The resolution of the optimum set of transformation parameters, at a courser level, is increased to be used as starting point for the next hierarchical level. Finally, after the application of the final deformation field a tensor reorientation is applied using the finite strain strategy.
$$\includegraphics[scale=1.25]{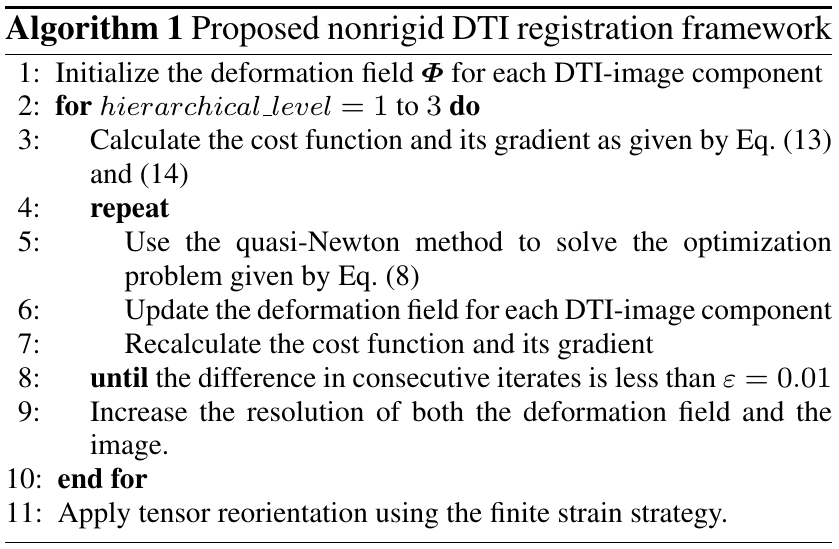}$$

\section{Experimental Results}
We tested the performance of the proposed approach on medical imaging data sets that were obtained from the National Alliance for Medical Image Computing (NAMIC) database in the MIDAS Journal~\cite{midas}. These data sets contain 20 cases: ten are normal controls and ten are Schizophrenic. Each data set contains images from several protocols, including T1-weighted (MR-T1), T2-weighted (MR-T2), diffusion weighted DWI, and fMRI. The images used in our experiments are DTI scans that were acquired on a $3$ Tesla GE system using an echo planar imaging (EPI) DTI Tensor sequence. The following scan parameters were used: $\text{TR}=17000 \text{ ms}$, $\text{TE}=78 \text{ ms}$, $\text{FOV}=24 \text{ cm}$, $144\times 144$ encoding steps, and $1.7 \text{ mm}$ slice thickness. The number of slices is 85 axial slices. In addition, $B_o$ field inhomogeneity maps are collected. To assess the registration accuracy of the proposed method on DTI data, we first applied a geometric distortion to a fixed image in order to generate a moving image. Then, we aligned the moving image with the fixed image. In all the experiments, we used the normalized histogram of the fixed image as the weight vector $\bg{\omega}$ in the multicomponent JT similarity measure. Moreover, the moving image is generated by applying a random perturbation to the corresponding fixed image using a thin-plate spline interpolation such that the mean nonrigid displacement of the pixels, caused by the relative displacement between the fixed and generated moving images, is the ground truth deformation field.
\subsection{Evaluation Criteria}
For the quantitative evaluation analysis, we considered only voxels having FA values larger than $0.4$. The registration accuracy of the proposed method is evaluated in terms of both the spatial registration and orientation correspondence.

\noindent\textbf{Deformation field correspondence:} The deformation field correspondence measure is defined as
\begin{equation}
\label{eq:C}
C_{B}=\frac{\Vert\bm{\Phi} - \bm{\Phi'}\Vert}{\Vert\bm{\Phi}\Vert+\Vert\bm{\Phi'}\Vert},
\end{equation}
where $C_B$ represents the distance between the estimated deformation field $\bm{\Phi'}$ and the ground-truth deformation $\bm{\Phi}$ for each voxel $B$. We then compute the median value, denoted by $C$, of $C_{B}$ for all selected voxels. The median value $C$ represents an overall measure of the deformation field correspondence, and takes values between $0$, when the estimated deformation field exactly matches ground truth deformation, and $1$ resulting in the worst alignment.

\noindent\textbf{First eigenvector angle difference:} To evaluate the quality of registration method with respect to the orientation information, the angle $a_{B}$ between the first eigenvector $\bm{n}_B$ of the fixed image and the deformed moving image $\bm{n}'_B$ can be calculated for each selected white matter voxel $B$ as follows:
\begin{equation}
\label{eq:aaa}
a_{B}=\cos^{-1} \left(\frac{\langle\bm{n}'_B,\bm{n}_B\rangle}{||\bm{n}'_B||\,||\bm{n}_B||}\right).
\end{equation}
The median, denoted by $a$, of all selected voxels $B$ is a measurement of preservation of orientation information after registration. The smaller the value of $a$, the better the orientation alignment between the involved images.

\noindent\textbf{Overlap of eigenvalue-eigenvector pairs:} The overlap of eigenvalue-eigenvector pairs between tensors is another measure of registration quality given by
\begin{equation}
\label{eq:ovl}
OVL =
\frac{1}{N_{B}}\sum_{B}\frac{\sum_{i=1}^{3}\lambda'_{i}\lambda_{i}
\langle\bm{\varepsilon}'_{i},\bm{\varepsilon}_{i}\rangle^2}{\sum_{i=1}^{3}
\lambda'_{i}\lambda_{i}},
\end{equation}
where $N_B$ is the total number of selected WM voxels, and $\lambda'_{i}$, $\lambda_{i}$, $\bm{\varepsilon}'_{i}$ and $\bm{\varepsilon}_{i}$ are eigenvalues and eigenvectors of the deformed moving image and fixed image, respectively. The maximum value $1$ of $OVL$ indicates complete overlap, whereas the minimum value $0$ represents no
overlap of the principal axes of the DT field.

\subsection{Qualitative Test}
In the first experiment, we distorted the fixed DW images of the data set shown in Figure~\ref{fig:test1} with a known nonrigid transformation field or the so-called ground truth deformation $\bm{\Phi}$, as shown in Figure~\ref{fig:test2}(a). Then, the DT field is computed from the deformed DW images. Next, the DT elements are
reoriented to preserve the alignment with the underlying, deformed microstructure. And then, the DW images are recomputed from the reoriented DT field, resulting in the moving image data set, as shown in Figure~\ref{fig:test2}(b)-(c). Next, we applied the proposed approaches using DT elements (JT-DT) and FA images (JT-FA), as well as the affine registration method based on mutual information~\cite{Leemans:05,Hecke:07}. Finally, we compared the registered DT and FA images to their corresponding fixed images. Figure~\ref{fig:test2} shows the results obtained from this experiment. It should be noted that the registered moving images obtained by JT-DT and JT-FA are visually more similar in shape to the fixed images than the images produced by the affine method. Moreover, it can be seen that the registered image using the affine registration method still has a considerable amount of misregistration. Unlike the affine method, most of the visible amount of misalignment in the moving image has been removed after applying our JT-DT and JT-FA approaches.

\begin{figure}[htbp]
\setlength{\tabcolsep}{.01em}
\centering
\begin{tabular}{ccc}
\includegraphics[width=1in,height=1.2in]{Deformation_Field.pdf}&
\includegraphics[width=1in,height=1.2in]{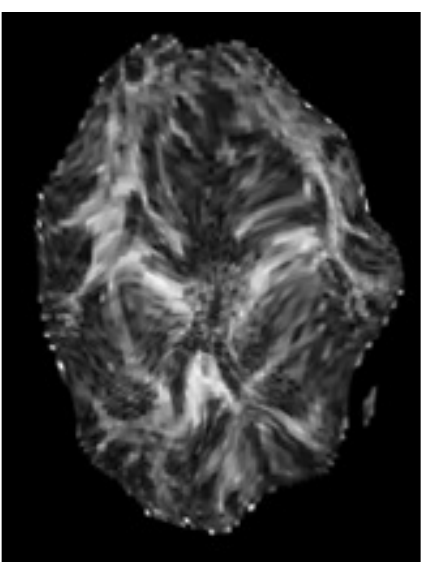}&
\includegraphics[width=1in,height=1.2in]{Moving_DTI.pdf}\\
(a) & (b) & (c) \\
\includegraphics[width=1in,height=1.2in]{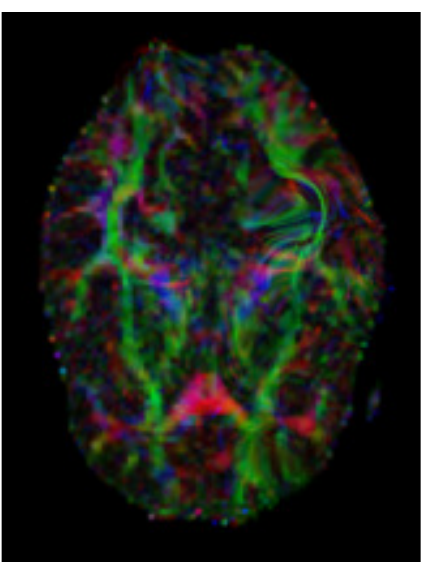}&
\includegraphics[width=1in,height=1.2in]{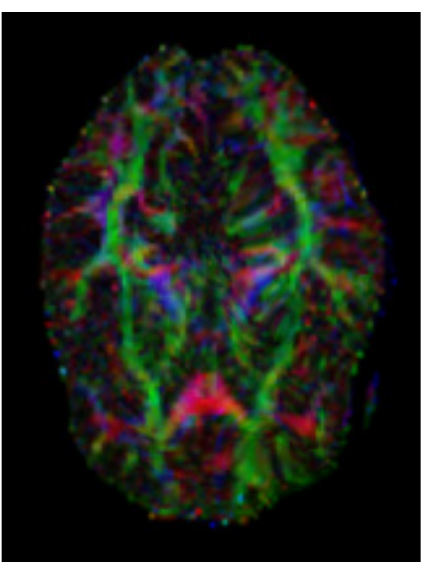}&
\includegraphics[width=1in,height=1.2in]{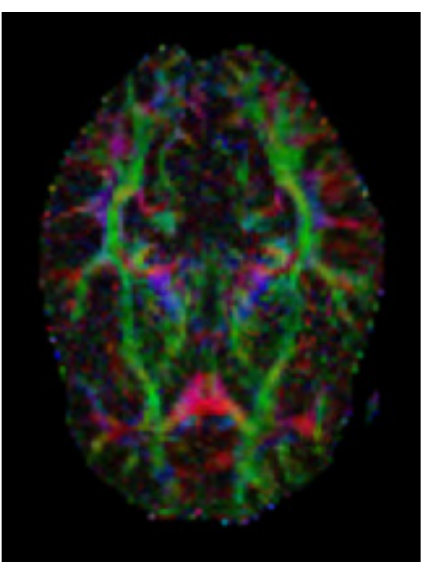}\\
(d) & (e) & (f)\\
\includegraphics[width=1in,height=1.2in]{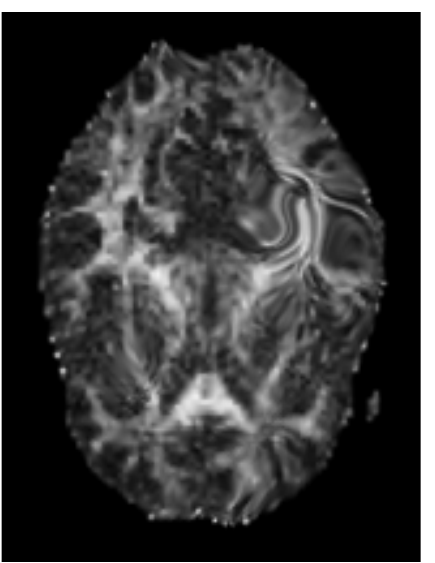}&
\includegraphics[width=1in,height=1.2in]{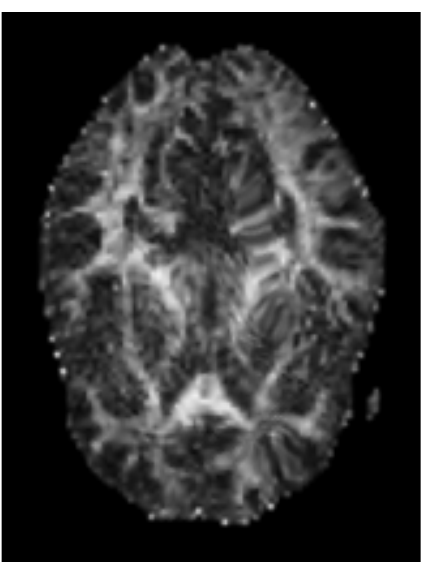}&
\includegraphics[width=1in,height=1.2in]{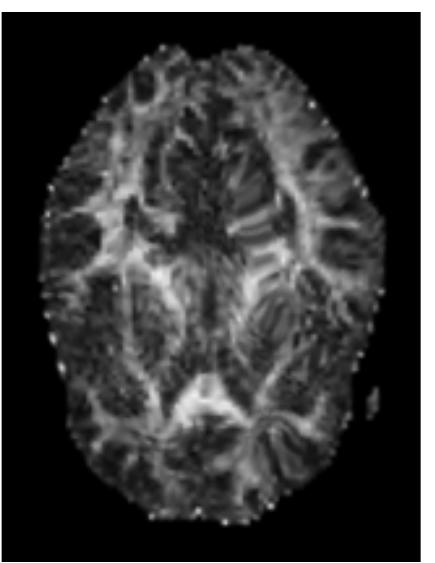}\\
(g) & (h) & (i)
\end{tabular}
\caption{Geometric distortion experiment: (a) Ground truth deformation field; (b)-(c) distorted FA and DT elements' images, respectively, with geometric distortion; (d)-(f) DT elements of the registered images using affine, JT-FA and JT-DT, respectively; (g)-(i) FA of the registered images using affine, JT-FA and JT-DT, respectively.} \label{fig:test2}
\end{figure}

\subsection{Quantitative Test}
In the second experiment, quantitative registration results on DTI data sets deformed with known deformation fields are shown in Figure~\ref{fig:testQuant}. All methods apply the finite strain strategy to reorient the tensors after
registration. In Figure~\ref{fig:testQuant}(a)-(b), the eigenvalue-eigenvector overlap ($OVL$) of tensors in corresponding voxels, and the first eigenvector angle difference ($a$) between the first eigenvectors of corresponding voxels are displayed. As shown in Figure~\ref{fig:testQuant}(a)-(b), the proposed registration approaches outperform the affine registration method. Moreover, the use of diffusion elements in the JT-DT method resulted in improved registration results compared to the JT-FA method. In Figure~\ref{fig:testQuant}(c), the measure $C$ calculates the discrepancy between the estimated deformation field and ground truth deformation field. The results obtained using the JT-DT method are considerably small compared to JT-FA. On the other hand, a paired $t$-test is used to determine if the difference in the quantitative parameters for the pairs of registration methods is statistically significant. The table displayed in Figure~\ref{fig:testQuant}(d) shows that at $95\%$ level of confidence, the JT-DT method significantly improves the registration accuracy compared to JT-FA and affine methods.

\begin{figure*}[htbp]
\setlength{\tabcolsep}{.5em}
\centering
\begin{tabular}{cc}
\includegraphics[scale=.45]{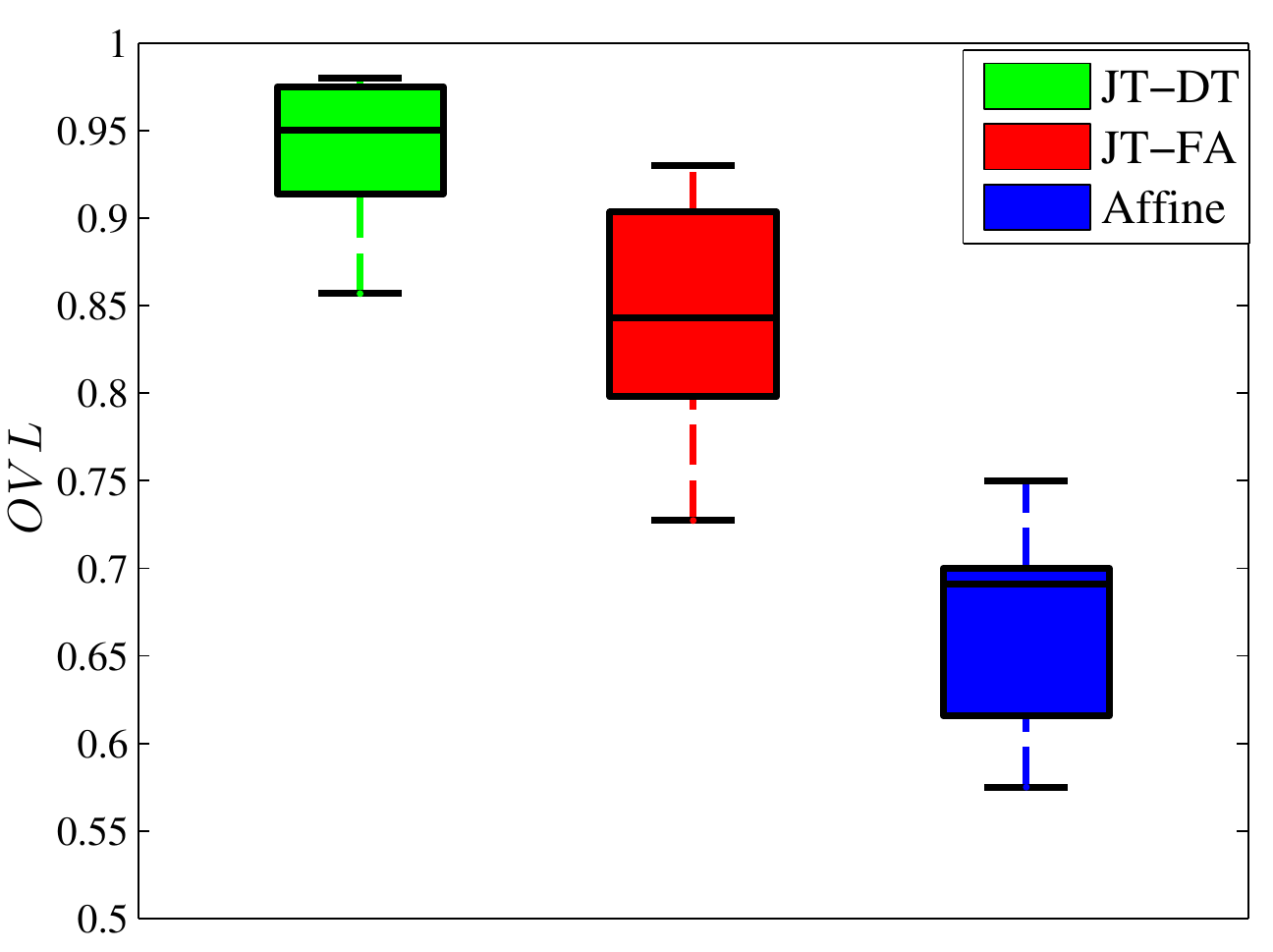} &
\includegraphics[scale=.45]{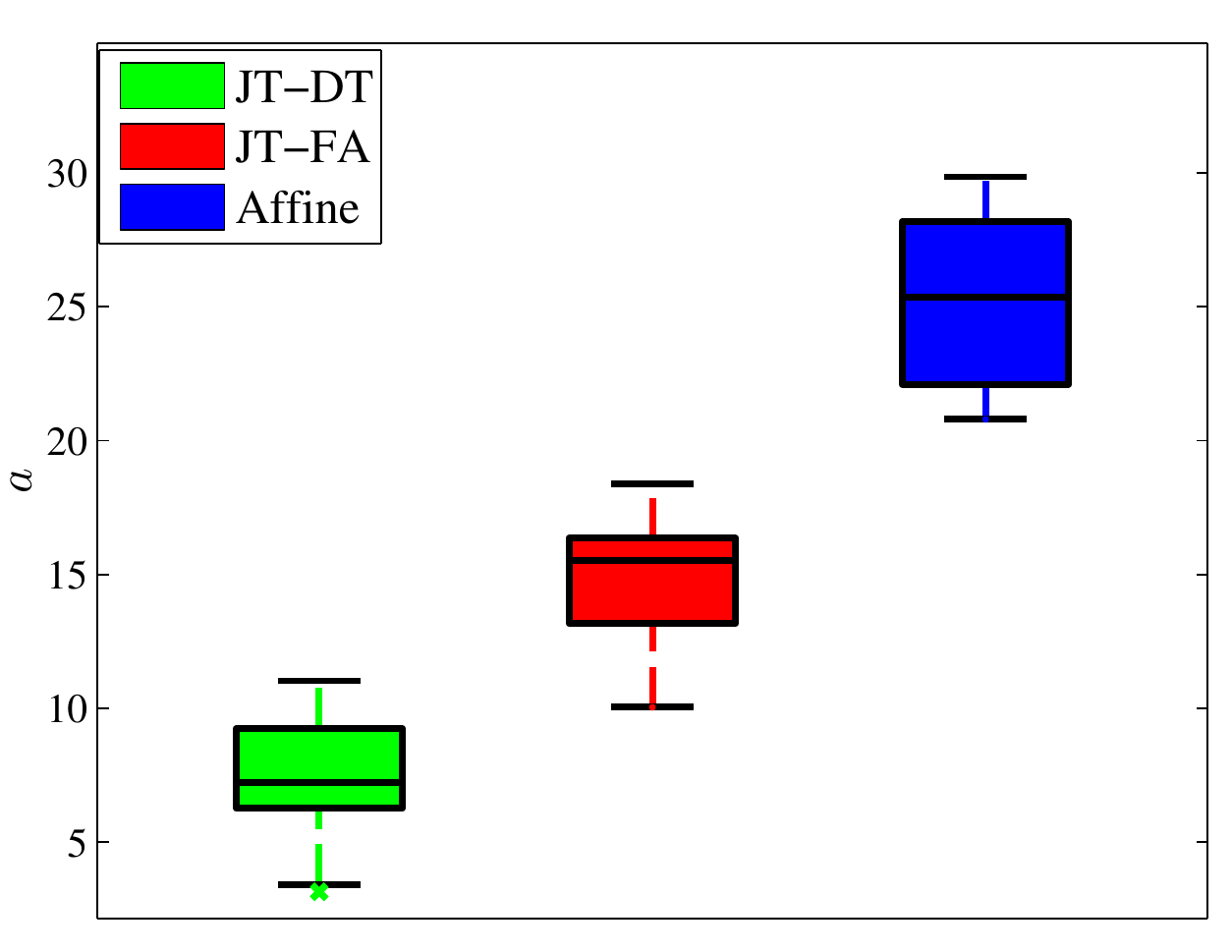} \\
(a) & (b) \\
\includegraphics[scale=.45]{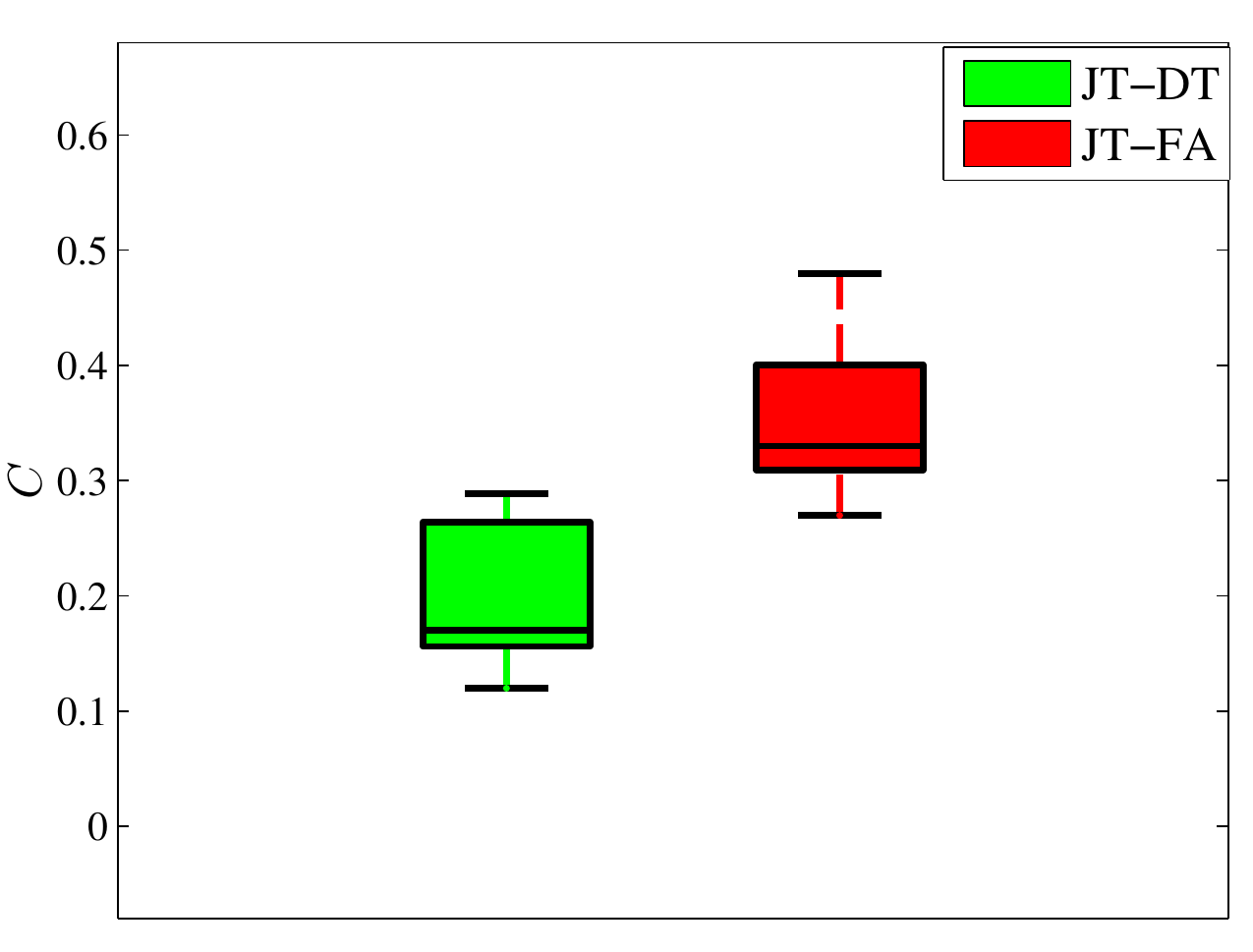}&
\includegraphics[scale=.8]{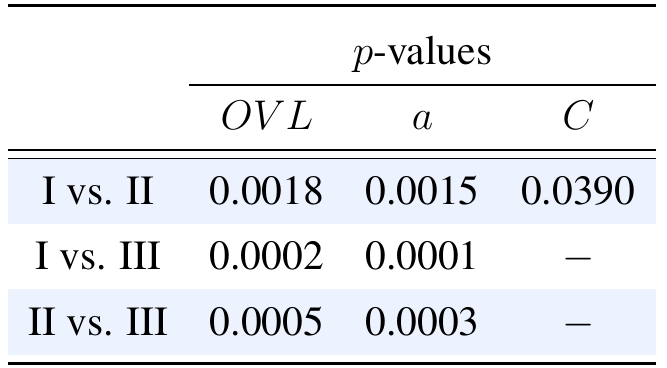} \\
(c) & (d)
\end{tabular}
\caption{Quantitative registration results of deformed image data sets with known deformation fields. JT-DT, JT-FA and affine registration methods are displayed on the horizontal axis, respectively. (a) The OVL measures the eigenvalue-eigenvector overlap of tensors in corresponding voxels between the fixed and registered images. (b) The angle difference between the first eigenvectors of corresponding voxels in the fixed and registered images. (c) The measure $C$ calculates the discrepancy between between the estimated and ground truth deformation fields. The $p$-values between the JT-DT, JT-FA and affine registration techniques represented by I, II and III, respectively, are displayed in the table.}
\label{fig:testQuant}
\end{figure*}

\section{Conclusions}
We proposed an information-theoretic technique for nonrigid registration of diffusion tensor images using a multicomponent similarity measure. In the proposed approach, we enabled explicit orientation optimization by incorporating tensor reorientation, which is necessary for wrapping DT images. The experimental results on DTI registration indicated the feasibility of the proposed approach and a much improved performance compared to the affine registration method.


\begin{thebibliography}{99}
\bibitem{Basser:94}  P. J. Basser, J. Mattiello, and D. Le Bihan, ``MR diffusion tensor spectroscopy and imaging,'' {\em Biophysical Journal}, vol. 66, no. 1, pp.
259-267, 1994.

\bibitem{Filippi:05}  M. Rovaris, A. Gass, R. Bammer, S. Hickman, O. Ciccarelli, D. Miller, and M. Filippi, ``Diffusion MRI in multiple sclerosis,'' {\em Neurology}, vol. 65, no. 10, pp. 1526–1532, 2005.

\bibitem{Alexander:01}  D.C. Alexander, C. Pierpaoli, P.J. Basser, and J.C. Gee, ``Spatial transformations of diffusion tensor magnetic resonance images,'' {\em IEEE Trans. Medical Imaging}, vol. 20, no. 11, pp. 1131–1139, 2001.

\bibitem{Jones:02} D. Jones, L. Griffin, D. Alexander, M. Catani, M. Horsfield, R. Howard, and S. Williams, ``Spatial normalization and averaging of diffusion tensor MRI data sets,'' {\em NeuroImage}, vol. 17, no. 2, pp. 592–617, 2002.

\bibitem{Guimond:02} A. Guimond, C. Guttmann, S. Warfield, and C. Westin, ``Deformable registration of DT-MRI data based on transformation invariant tensor characteristics,'' {\em Proc. ISBI}, pp. 1-4, 2002.

\bibitem{Xu:03} D. Xu, S. Mori, D. Shen, P. van Zijl, and C. Davatzikos, ``Spatial normalization of diffusion tensor fields,'' {\em Magnetic Resonance in Medicine}, vol. 50, no. 1, pp. 175-182, 2003.

\bibitem{Kikinis:02} J. Ruiz-Alzola, C-F. Westin, S. Warfield, A. Nabavi, and R. Kikinis, ``Nonrigid registration of 3D tensor medical data,'' {\em Medical Image Analysis}, vol. 6, no. 2, pp. 143-161, 2002.

\bibitem{Alexander:99} D. Alexander, J. Gee, and R. Bajcsy, ``Elastic matching of diffusion tensor MRIs,'' {\em Proc. CVPR}, pp. 244-249, 1999.

\bibitem{Zhang:06} H. Zhang, P. Yushkevich, D. Alexander, and J. Gee, ``Deformable registration of diffusion tensor MR images with explicit orientation optimization,'' {\em Medical Image Analysis}, vol. 10, no. 5, pp. 764-785, 2006.

\bibitem{Younes:06} Y. Cao, M. Miller, S. Mori,R. Winslow, and L. Younes, ``Diffeomorphic matching of diffusion tensor images,'' {\em Proc. CVPR}, 2006.

\bibitem{Hecke:07} W. Hecke, A. Leemans, E. D'Agostino, S. De Backer, E. Vandervliet, P.M. Parizel, and J. Sijbers ``Affine coregistration of diffusion tensor magnetic resonance images using mutual information,'' {\em IEEE Trans. Medical Imaging}, vol. 26, no. 11, pp. 1598-1612, 2007.

\bibitem{Westin:03} H.J. Park, M. Kubicki, M. Shenton, A. Guimond, R. McCarley, S. Maier, R. Kikinis, F. Jolesz, and C.F. Westin, ``Spatial normalization of diffusion tensor MRI using multiple channels,'' {\em NeuroImage}, vol. 20, no. 4, pp. 1995-2009, 2003.

\bibitem{Thirion:98} J.P. Thirion, ``Image matching as a diffusion process: An analogy with maxwell's demons,'' {\em Medical Image Analysis}, vol. 2, no. 3, pp. 243-260, 1998.

\bibitem{Anderson:01} A. Anderson, ``Theoretical analysis of the effects of noise on diffusion tensor imaging,'' {\em Magnetic Resonance in Medicine}, vol. 46, no. 6, pp. 1174-1188, 2001.

\bibitem{Hamza:03} A. Ben Hamza and H. Krim, ``Jensen-R\'enyi divergence measure: theoretical and computational perspectives,'' {\em Proc. IEEE Int. Symp. Information Theory}, 2003.

\bibitem{Khader:11} M. Khader and A. Ben Hamza, ``Non-rigid image registration using an entropic similarity,'' {\em IEEE Trans. Information Technology in Biomedicine}, vol. 15, no. 5, pp. 681-690, 2011.

\bibitem{Khader:12} M. Khader and A. Ben Hamza, ``An information-theoretic method for multimodality medical image registration,'' {\em Expert Systems with Applications}, vol. 39, no. 5, pp. 5548-5556, 2012.

\bibitem{Nocedal:02} J. Nocedal and S. J. Wright, {\em Numerical Optimization}. New York: Springer-Verlag, ch. 8-9, 2000.

\bibitem{Leemans:05} A. Leemans, J. Sijbers, S. De Backer, E. Vandervliet, and P.M. Parizel ``Nonrigid coregistration of diffusion tensor images using a viscous fluid model and mutual information,'' {\em Proc. ACIVS}, LNCS 3708, pp. 523–530, 2005.

\bibitem{Taylor:04} W.D. Taylor, E. Hsu, K.R. Krishnan, and J.R. MacFall, ``Diffusion tensor imaging: background, potential, and utility in psychiatric research,'' {\em Biological Psychiatry}, vol. 55, no. 3, pp. 201-207, 2004.

\bibitem{Mori:95} S. Mori and P. Zijl, ``Diffusion weighting by the trace of the diffusion tensor within a single scan,'' {\em Magnetic Resonance in Medicine}, vol. 33, pp. 41-52, 1995.

\bibitem{Tsallis:88} C. Tsallis, ``Possible generalization of Boltzmann-Gibbs statistics,'' {\em Journal of Statistical Physics}, vol. 52, pp. 479-487, 1988.

\bibitem{Mattes:03} D. Mattes, D. R. Haynor, H. Vesselle, T. K. Lewellen, and W. Eubank, ``PET-CT image registration in the chest using free-form deformations,'' {\em IEEE Trans. Medical Imaging}, vol. 22, no. 1, pp. 120-128, 2003.

\bibitem{midas}[Online]. http://insight-journal.org/midas/collection/view/190.

\bibitem{Simmons:99} D. Jones, M. Horsfield, and A. Simmons, ``Optimal strategies for measuring diffusion in anisotropic systems by magnetic resonance imaging,'' {\em Journal of Magnetic Resonance in Medicine}, vol. 42, no. 3, pp 515-525, 1999.
\end{thebibliography}
\end{document}